\pdfoutput=1

\documentclass[11pt]{article}

\usepackage[final]{acl}

\usepackage{times}
\usepackage{latexsym}

\usepackage{custom}
\usepackage{supertabular}
\usepackage{booktabs}
\usepackage{caption}
\usepackage{array}
\usepackage{tcolorbox}
\usepackage{xcolor}
\definecolor{DiverseMagenta}{RGB}{255, 0, 128}
\definecolor{AccentBlue}{RGB}{0, 128, 255}

\usepackage{tabularx}
\usepackage{CJKutf8}
\usepackage{listings}

\usepackage[T1]{fontenc}

\usepackage[utf8]{inputenc}

\usepackage{microtype}

\usepackage{inconsolata}

\usepackage{graphicx}
\usepackage{makecell}

\makeatletter
\AddToHook{cmd/x@supertabular/after}
   {%
     \let\orig@startpbox@action\@startpbox@action
     \let\@startpbox@action\@startpbox
   }
\AddToHook{cmd/ST@restore/after}
   {%
     \let\@startpbox@action\orig@startpbox@action
   }
\makeatother

\title{TactfulToM: Do LLMs Have the Theory of Mind Ability \\ to Understand White
Lies?}

\author{
Yiwei Liu\textsuperscript{$\heartsuit$}\thanks{Work conducted during an internship at National Institute of Informatics.}
\quad
Emma Jane Pretty\textsuperscript{$\spadesuit$}\thanks{Contributed equally as second authors.} \quad
Jiahao Huang\textsuperscript{$\diamondsuit$}\footnotemark[2] \quad
Saku Sugawara\textsuperscript{$\clubsuit$} \\
\\
\textsuperscript{$\heartsuit$} EPFL, Lausanne, Switzerland \quad
\textsuperscript{$\spadesuit$} Tampere University \quad \\
\textsuperscript{$\diamondsuit$} University of Tokyo \quad
\textsuperscript{$\clubsuit$} National Institute of Informatics \\
\texttt{yiw.liu@epfl.ch saku@nii.ac.jp}
}

\begin{document}
\maketitle
\begin{abstract}
While recent studies explore Large Language Models' (LLMs) performance on Theory of Mind (ToM) reasoning tasks, research on ToM abilities that require more nuanced social context is limited, such as white lies. We introduce TactfulToM, a novel English benchmark designed to evaluate LLMs' ability to understand white lies within real-life conversations and reason about prosocial motivations behind them, particularly when they are used to spare others' feelings and maintain social harmony. 
Our benchmark is generated through a multi-stage human-in-the-loop pipeline where LLMs expand manually designed seed stories into conversations to maintain the information asymmetry between participants necessary for authentic white lies. We show that TactfulToM is challenging for state-of-the-art models, which perform substantially below humans, revealing shortcomings in their ability to fully comprehend the ToM reasoning that enables true understanding of white lies.\footnote{https://github.com/nii-cl/tactful-tom}
\end{abstract}

\section{Introduction}
\begin{figure}[ht]
    \raggedright 
    \centering
    \includegraphics[width=1\linewidth]{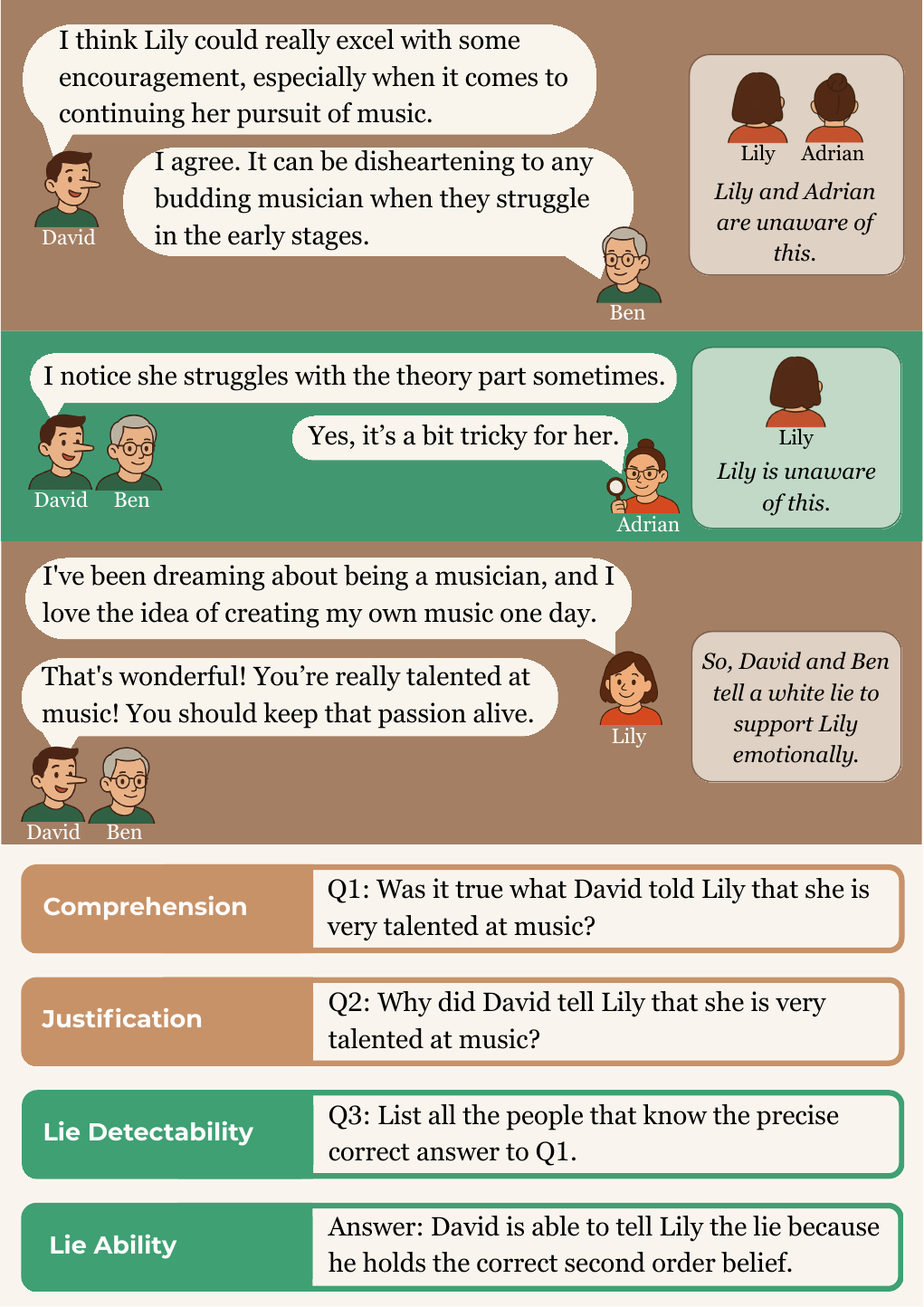}
    \caption{Example from a question set in TactfulToM.}
    \label{fig:example}
\end{figure}

Theory of Mind (ToM) is the cognitive ability to impute mental states to oneself and others, and to use these inferred mental representations to predict and explain behaviors~\cite{premack1978does, baron1985does}. This ability is recognized as a foundation for effective social interactions and a pillar of common sense reasoning~\cite{lake2017building}, which is crucial for developing human-level AI systems. Modern LLMs like GPT \cite{hurst2024gpt} and DeepSeek \cite{deepseekai2025deepseekr1incentivizingreasoningcapability} have demonstrated remarkable reasoning capabilities in structured domains such as mathematics and programming, yet research consistently reveals significant gaps between humans and LLMs in ToM tasks, especially when applied to realistic social scenarios~\cite{chen-etal-2024-tombench, gu2024simpletom}. 

Among the various sub-abilities of ToM, understanding white lies, intentional falsehoods told specifically to protect others' feelings, represents a particularly complex aspect that combines belief tracking with emotional sensitivity~\cite{beaudoin2020systematic, abdollahi2022artificial}. The ability to detect white lies and understand their emotional motivations becomes essential for developing safe and appropriately responsive AI tools, especially as LLM tools are increasingly deployed in domains requiring emotional intelligence, such as educational tutoring, medical consultation, and caregiving. Despite this importance, white lies remain largely understudied.

ToM battery evaluations like ToMBench~\cite{chen-etal-2024-tombench} have included white lie tasks but with limitations, only containing 20 white lie samples without dialogue interaction. Testing on such small samples is insufficient for reliable evaluation, as minor variations in test cases can significantly alter results~\cite{ullman2023largelanguagemodelsfail}.
Additionally, using established psychological ToM tests risks data contamination that could artificially inflate performance metrics~\cite{shapira2023clever, chen-etal-2024-tombench}. This creates a critical research gap in understanding LLMs' white lie comprehension capabilities despite the significance for AI systems to safely operate in nuanced contexts.

To address this challenge, we introduce TactfulToM, an English benchmark that aims to evaluate LLMs' ability to understand and reason about white lies in real-world conversational contexts, particularly focusing on the interplay between deceptive statements and their underlying motivations. Our benchmark offers four key contributions: (1) a novel decomposition framework that breaks down white lies into triplets and role-based information asymmetry, enabling manually crafted seed stories; (2) high-quality conversations generated via human-in-the-loop generation pipeline (avoiding biases from direct LLM generation) with strict validation; (3) a comprehensive evaluation framework to test models' understanding of white lies by combining mental state tracking questions with both established measures from Strange Stories~\cite{happe1994advanced} and our newly designed question types; and (4) a diverse dataset of 100 multi-party conversations spanning across different white lie classes, types, and (falsifiability) difficulty levels, which contains 6.7K questions across multiple answer formats.

We evaluate TactfulToM on nine recent LLMs from four different families, including both vanilla and reasoning models. Through our experiments, we uncover gaps between human and AI performance in white lie comprehension. The analysis of evaluation results on TactfulToM reveals several interesting findings: (1) all tested LLMs significantly underperform humans, even the best-performing ones (DeepSeek families and GPT-4o); (2) Chain-of-Thought (CoT) prompting and specialized reasoning models show inconsistent improvements, with some models even performing worse than vanilla models from the same families; (3) LLMs struggle with true white lie understanding and fail to grasp the genuine motivations behind white lies; and (4) LLMs can track mental states but fail to apply them effectively in white lie contexts. 

Our contributions are summarized as follows:
\begin{itemize}
\item We present a benchmark that tests LLMs' ability to understand white lies in social contexts, filling a research gap in ToM evaluation.
\item Our dataset covers five white lie classes, two types, and three levels, all constructed efficiently using a human-in-the-loop process.
\item Our analysis reveals limitations in the white lie reasoning capabilities of recent LLMs, providing insights for future model development.
\end{itemize} 

\section{TactfulToM Design}
\begin{figure*}[!t]
    \centering
    \includegraphics[width=\textwidth]{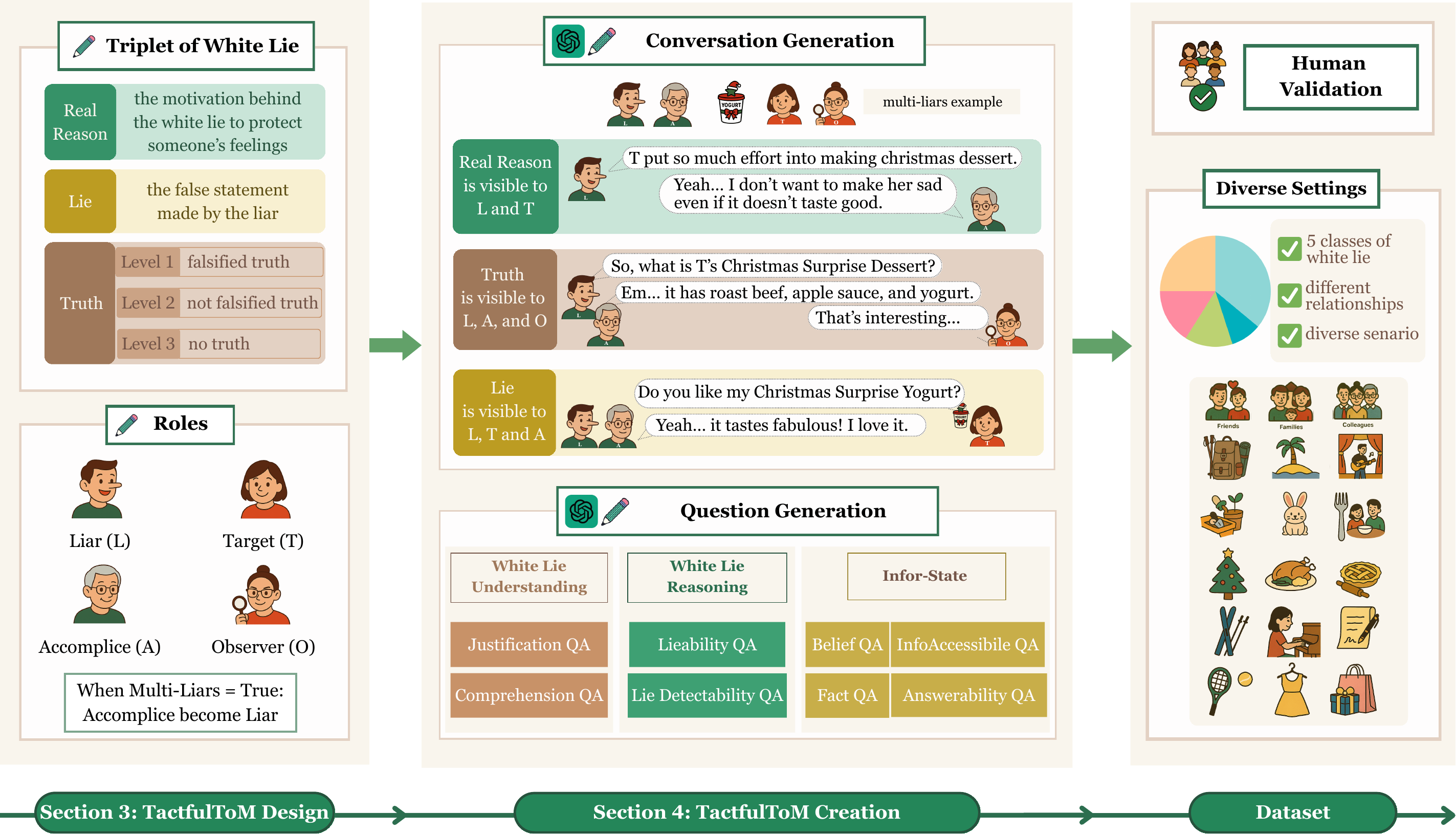}
    \caption{Overview of the dataset construction pipeline for TactfulToM.}
    \label{fig:pipeline}
\end{figure*}

Building upon the white lie test from Strange Stories~\cite{happe1994advanced} and previous successful evaluations of LLMs' ToM ability~\cite{kim-etal-2023-fantom}, we developed a dataset of social conversations capturing common white lies in daily life. This section outlines our design considerations and approach (as shown in Figure~\ref{fig:pipeline}): (1) theoretical requirements informing our design; (2) methodology for structuring white lies with triple and role-based information asymmetric; and (3) evaluation framework for white lie understanding and reasoning.

\subsection{Theoretic Requirements from ToM Task Designing}
\label{subsec:theoretic-requirement}
ToM evaluation requires carefully structured scenarios that test a model's ability to accurately attribute mental states. Three critical aspects based on \citet{quesque2020theory,kim-etal-2023-fantom} are identified: Non-merging Mental States, Non-mentalising, and Elimination of Visual Indicators.

\paragraph{Non-merging Mental States} 
A valid evaluation of ToM requires the model to distinguish between its own knowledge and the beliefs of others. In scenarios where one character provides false information and others either believe the lie or know the truth, the model must infer what a deceived character believes only based on the information available to them, not based on the model's knowledge. To ensure the non-merging requirement, scenarios must involve multiparty conversations where it is explicitly revealed who knows the truth and the lie. This allows for controlled belief divergence, ensuring that the model must track the different perspectives of each character rather than assuming all characters share the same understanding. We design our benchmark with information asymmetry to enforce this differentiation.

\paragraph{Non-mentalising}
It is crucial not to attribute model success to genuine mentalizing when simpler processes can explain the outcome. In white lie scenarios, if a model’s correct answer arises from surface-level patterns or word correlations, this explanation should be prioritized over more complex reasoning about mental states. For example, the model might correctly identify that a character believes a lie, but if this answer is based on word associations rather than true mental state reasoning, it suggests the model is not engaging with the belief system of the character. To address this, we introduce distractor answers with high word correlation to test if the model is relying on deeper reasoning rather than simple associations.

\paragraph{Elimination of Visual Indicators}
The model should also not rely on descriptions of body language, emotions, or visual indicators when inferring belief states, only linguistic contexts \cite{premack1978does, baron1985does}. Relying on such cues would lead to shortcuts that allow the model to infer beliefs based on visible indicators, not through genuine reasoning about what another person might believe. Thus, our benchmark contains conversational exchanges without any narrative descriptions, requiring the model to infer mental states purely from the dialogue, ensuring that belief inference is based on logical reasoning rather than perceptual cues.

\subsection{Structuring White Lies}
\label{subsec:structuring-white-lies}
\paragraph{White Lie Triplet Decomposition} 
To systematically create our dataset, we first decompose white lies into three elements: (1) \textbf{Real Reason}: the motivation behind telling the lie; (2) \textbf{Lie}: the false statement made by the liar; (3) \textbf{Truth}: the actual truth that diverges from the lie. For example, in a classic Strange Story test~\cite{happe1994advanced}, the truth is ``Helen wanted a rabbit but received encyclopedias from her parents'' while Helen lies ``It's lovely, thank you. It's just what I wanted.'' with the real-reason being to avoid hurting Helen's parents' feelings after they gave her a gift they thought she would like.

\paragraph{Real-Reason Correspondence to White Lie Types}
White lies fall into two distinct types based on their underlying motivations: altruistic white lies and Pareto white lies ~\cite{erat2012white}. Altruistic white lies are told purely for the benefit of others, where the liar may incur some personal cost or disadvantage. In contrast, Pareto white lies create a mutually beneficial outcome, serving both the interests of the person being lied to and the liar themselves. The fundamental categorization guided our design of two types of real-reason statements corresponding to these two categories of lies. 

\paragraph{Three Levels of Truth Accessibility in White Lies}
To reflect real-world complexities, we incorporate three difficulty levels by varying falsifiability (between ``lie'' and ``truth'' ) in our white lie triplets. After establishing the ``real reason'' (e.g., ``declining an invitation without hurting T's feelings'') and ``lie'' (e.g., ``L has a reservation tonight''), we determine how the truth is presented. We structure conversations into three difficulty levels: (1) Level-1: falsifiable truth provided, e.g., ``L does not have a reservation tonight''; (2) Level-2: non-falsifiable truth provided, e.g., ``L hasn't decided what to do tonight''; and (3) Level-3: no truth provided. Not all white lie scenarios can reasonably accommodate all three levels; some contexts intrinsically require truth disclosure while others cannot reasonably support ambiguous truth construction. As such, we selectively designed appropriate levels for each white lie triplet. This creates progressive reasoning challenges: with the truth provided, models can identify lies before determining the motivation; without it, models must infer the deceptive nature directly from the real reason, thus compelling models to rely on conversational context and reasoning rather than simply detecting information contradictions to reverse-engineer the scenarios.

\paragraph{Role-based Information Asymmetry}
Building upon the inherent characteristics of white lie scenarios, we define four roles based on their access to the white lie triplet: the \textbf{Liar (L)}, who has complete understanding and knowledge of the white lie; the \textbf{Accomplice (A)}, who has access to all elements in triplet; the \textbf{Observer (O)}, who only knows the truth; and the \textbf{Target (T)}, who only receives the lie. This asymmetric access leads to varying degrees of white lie comprehension among participants; it is not only a necessary condition for white lies to exist, but also aligns with the non-merging mental states requirement (Section~\ref{subsec:theoretic-requirement}). Our dataset incorporates diverse character relationships (friends, families, and colleagues) and complex patterns including multi-liar scenarios where accomplices function as additional liars given their equivalent information access. This design captures more diverse and realistic interaction patterns. We also impose a crucial constraint: all discussions about the white lie triplet begin within the conversation scenario, with no prior exchange of this information among characters.

\subsection{Hierarchical Evaluation Framework: Mental States to White Lie Reasoning}
\label{subsec:question-category}
Our evaluation framework employs a progressive three-tier question hierarchy: \textbf{(A) Info-State Questions} assesses basic mental state tracking, \textbf{(B) First-Order: White Lie Understanding} evaluates how models perceive and interpret white lies, and \textbf{(C) Second-Order: White Lie Reasoning} tests the models' ability to reason about different roles' perspectives on the white lie within the conversation. 

\paragraph{Info-State Questions} 
We include four question types targeting belief attribution: first, we establish \textbf{Fact} questions (factQ) that include factual question-answer pairs about the asymmetrical information ``real reason'' and ``truth''. Building on these, we develop \textbf{Belief} questions that assess first-order beliefs (what characters believe) and second-order beliefs (how characters understand others' beliefs: ``What does X believe about Y's understanding of [FactQ]?''). We also include \textbf{Info Accessibility} questions (``List all characters who know [real reason/truth]'') and \textbf{Answerability} questions (e.g., ``List all characters who can answer: [FactQ]''). This question-type structure prevents inflated scores from the ``illusion of ToM''~\cite{kim-etal-2023-fantom} while enabling more accurate assessment of mental state tracking capabilities.

\paragraph{White Lie Understanding (1st-Order)}
Drawing from the Strange Story test, we assess basic white lie understanding through two question types: comprehension and justification. \textbf{Comprehension} questions (``Is the statement X told Y true?'') evaluate whether models can identify false statements as lies. \textbf{Justification} questions (``Why did X say that to Y?'') probe whether models recognize the prosocial motivations behind white lies that distinguish a white lie from a simple deception. These questions are complementary; even if a model correctly identifies a statement as false, it must also understand the protective intention to fully comprehend the white lie concept. According to \citet{happe1994advanced}, accurate responses to both questions indicate second-order ToM ability.

\paragraph{White Lie Reasoning (2nd-Order)}
We introduce two novel question types that evaluate models' understanding of characters' perspectives: \textbf{Lie Ability} questions if models can identify which characters possess the necessary conditions to tell a white lie (requiring understanding Liar's second-order beliefs). \textbf{Lie Detectability} questions evaluate if models can determine which characters have sufficient information to recognize deception. Both require reasoning about characters' information access and resulting beliefs, providing a more stringent test of genuine second-order ToM reasoning beyond simple pattern matching.

\paragraph{Comprehensive Evaluation Format}
To ensure robust evaluation, we present each question in two formats. Multiple-choice questions (MCQs) are complemented with free-form responses to assess genuine understanding in addition to choice selection, as providing choices inherently guides model reasoning paths. Similarly, list-type questions are presented in both open-ended and binary formats. This comprehensive approach allows us to check cases where providing answer options may guide the model to reason.

\paragraph{Design Consideration for Evaluation of Emotion-based Real Reasons}
When the real reason involves protecting the target from emotional distress unrelated to the lie itself (e.g., "Lynn will be very sad if she knows her grandma passed away"), our evaluation framework requires adaptation. Unlike factual information that can be explicitly discussed within conversations, emotional reactions (such as sadness) are inherent to the target and cannot be treated as discrete information. We exclude Information Accessibility and Lie Ability questions for such cases, which applies specifically to our {common sense (emotional soothing)} (Class 2) category (Section~\ref{sec:conversation generation}), ensuring framework consistency while accommodating emotion-based white lies.

\section{TactfulToM Creation}
The construction of TactfulToM consists of the following steps (as shown in Figure~\ref{fig:pipeline}): (1) manually creating seed stories, and then expanding them into natural conversations through a human-in-the-loop process; (2) generating question-answer pairs through templates; and (3) strict quality control. 
\subsection{Conversation Generation}
\label{sec:conversation generation}
The construction of TactfulToM consists of the following steps: (1) manually create seed stories, and then (2) expand them into natural conversations through a human-in-the-loop process.
\paragraph{Seed Stories} To create dataset diversity, we collected examples from interviews, social media, and online sources documenting white lie scenarios in daily life. We gathered examples in the format of white lie triplets to systematically capture the essential components of each scenario. We then categorized them into five distinct classes based on different motivations behind white lies: \textbf{social evasion} (Class 0), \textbf{common sense (imagination preservation)} (Class 1),  \textbf{common sense (emotional soothing)} (Class 2),  \textbf{confidence enhancement} (Class 3), and \textbf{mistake hiding} (Class 4). This categorization represents both altruistic white lies (Classes 1, 2, and 3) and Pareto white lies (Classes 0 and 4), ensuring comprehensive coverage of realistic social interactions. We constructed 100 seed stories, the data distribution across these categories is provided in Appendix~\ref{appendix:dataset_distribution}.

\paragraph{Generation Pipeline}
To facilitate conversation generation, we designed a set of scenario elements and combined them with seed stories as input for our 4-step generation prompt template, in which each step generates one element of the white lie triplet sequentially (conversation generation template is provided in Appendix~\ref{appendix:conversation prompt template}). The generation process employed GPT-4o~\cite{hurst2024gpt} in a human-in-the-loop methodology. Following FANToM~\cite{kim-etal-2023-fantom}, we assigned character names sampled from the top 20\% most popular US birth names~\cite{burnsworth2022names} and simulated scenarios where characters temporarily leave the conversation, allowing remaining participants to discuss without the absent participants' knowledge. We expanded the leaving reasons from~\citet{kim-etal-2023-fantom} into a more comprehensive list (samples provided in Appendix~\ref{appendix:leaving_reason}).

This stepwise generation pipeline and controlled asymmetry serve two primary purposes. First, it enables the model to generate white lie conversations even when its understanding of white lies is limited, by preventing the model from developing its own interpretation of white lies, thus avoiding conversations generated based on the model's potentially flawed understanding and reducing generation bias (see Appendix~\ref{appendix:clarification} for detailed explanation of how we address GPT-4o's white lie understanding limitations). Second, this approach enables controlled information asymmetry by systematically managing each participant's involvement at different conversation segments. where multiple dialogue segments were generated and carefully selected to ensure each segment contained only the intended information and roles before being combined into complete conversations. 

\subsection{Question-Answer Pair Generation}
\label{sec:qapair}
We developed a systematic templated generation approach for all question types (introduced in Section~\ref{subsec:question-category}), where templates are populated with white lie triplet elements and role information, enabling efficient question generation. All templates and examples are provided in Appendix~\ref{appendix:question_template}. Additionally, we systematically generated wrong options for MCQs to ensure each question has one correct answer and several high-quality but misleading distractors. For most question types, we automated this process using formalized operators, while justification questions required few-shot prompting to generate semantically diverse wrong options. Examples are provided in Appendix~\ref{appendix:Justification-options}.

\subsection{Strict Quality Control of TactfulToM}
\label{sec:validation}
We employed a multi-stage approach for strict quality control. For constructing seed stories, graduate students reviewed all white lie triplets to ensure logical consistency. During the generation, we created multiple versions (typically requiring 3-4 iterations) of each conversation segment and selected the best version to achieve natural dialogue flow while preserving the intended information asymmetries (more details can be found in Appendix~\ref{appendix:generation_quality}). 

For final validation, we recruited 21 annotators from the Prolific platform\footnote{\url{https://www.prolific.com/}} who met high-standard selection criteria. From an initial pool of 79 candidates, we conducted a qualification test designed to verify the ability to evaluate conversation coherence, resulting in the selection of 21 qualified annotators. Each conversation was reviewed by three independent annotators who flagged potential issues with coherence, safety, or white lie authenticity. While we received occasional flags, no conversation received majority votes for removal (detailed screening criteria, qualification test design, and validation results can be found in Appendix~\ref{appendix:annotation}).

\subsection{Statistics}
\label{Statistic for TactfulToM}
TactfulToM comprises 100 conversations spanning 5 white lie categories and 3 difficulty levels (based on falsifiability), containing 6.7K questions across multiple formats: multiple-choice, binary, and free-form questions. The dataset features an average of 23.66 tokens per question, with conversations averaging 32.6 turns and 33.0 tokens per turn. Table~\ref{tab:statistics_compare} presents a comparison of basic statistics between TactfulToM and other ToM benchmarks.
\begin{table}[H]
\centering
\resizebox{\columnwidth}{!}{%
\begin{tabular}{lcccc}
\toprule
\textbf{Dataset} & 
\textbf{\begin{tabular}[c]{@{}c@{}}Total \\ \#Convs\end{tabular}} & 
\textbf{\begin{tabular}[c]{@{}c@{}}Total \\ \#Questions\end{tabular}} & 
\textbf{\begin{tabular}[c]{@{}c@{}}Avg. Length \\ per Turn \\ (\#Tokens)\end{tabular}} & 
\textbf{\begin{tabular}[c]{@{}c@{}}Avg. \#Turns \\ per Conv.\end{tabular}} \\
\midrule
ToMATO & 753  & 5.4K & 41.6 & 16.0 \\
FANToM & 256  & 10K  & 31.4 & 24.5 \\
TactfulToM (Ours) & 100 & 6.7K & 33.0 & 32.6 \\
\bottomrule
\end{tabular}%
}
\caption{A comparison of dataset statistics between TactfulToM, ToMATO~\cite{shinoda2025tomato} and FANToM~\cite{kim-etal-2023-fantom}. Conv. refers to conversation.}
\label{tab:statistics_compare}
\end{table}

\section{Experiments}
\subsection{Model Choice}
We test nine LLMs from four families, including vanilla and reasoning models (indicated by $^*$):
    \textbf{GPT}: gpt-4o-2024-08-06~\cite{hurst2024gpt}, o1-2024-12-17$^*$~\cite{jaech2024openai}, o3-mini-2025-01-31$^*$\footnote{\url{https://openai.com/index/openai-o3-mini/}}; 
    \textbf{DeepSeek}: DeepSeek-V3-0324~\cite{deepseekai2024deepseekv3technicalreport}, DeepSeek-R1-Turbo$^*$~\cite{deepseekai2025deepseekr1incentivizingreasoningcapability};
    \textbf{Llama}: Llama-3.3-70B-Instruct~\cite{grattafiori2024llama};
    \textbf{Qwen}: Qwen2.5-72B-Instruct~\cite{yang2025qwen3}, QwQ-32B$^*$~\cite{qwq32b}. 

For the vanilla non-reasoning models, we employed two types of prompt templates: one that guided them to generate a direct answer, and another that guided them to generate the answer after a Chain-of-Thought (CoT) process. For the reasoning models, we only used the direct-answer prompt template. We presented the prompt templates in Table \ref{tab:prompt_template}, Appendix \ref{appendix:prompt template}. In addition to CoT, we also tested a compatible ToM-specific method (Shoes-of-Others prefixing~\cite{shinoda2025letssallysshoesshoesofothers}) for our white lie scenarios, with results in Appendix~\ref{appendix:tom_methods}.

\subsection{Metrics}
We employ four question formats across our evaluation framework: MCQs, binary, list-type, and free-form responses. Comprehension, Justification, Lie Ability, Belief, and Fact questions (except Lie Ability: MCQs only), while Lie Detectability, Info Accessibility, and Answerability questions use the binary and list formats. For structured responses (MCQs, binary, and list), we use accuracy as the primary evaluation metric and conduct detailed analyses of error patterns. For freeform responses, we determine the closest option using three complementary methods: cosine similarity (all-MiniLM-L6-v2\footnote{\url{https://huggingface.co/sentence-transformers/all-MiniLM-L6-v2}}), token-F1, and LLM-as-judge (DeepSeek-v3). Given the varying chance levels across formats, we report the MCQs and list format results, while using free-form responses for in-depth analysis.

\subsection{Human Performance} We evaluated human performance through 12 participants (9 annotators and 3 graduate students) on 15 sets of questions (drawn through stratified random sampling from the 100 question sets to ensure balanced representation across all five question categories and three complexity levels). To remove redundancy, we selected one format for each question type as follows: Comprehension [binary], Justification [MCQs], Lie Ability [MCQs], Lie detectability [list], belief [MCQs], Information Accessibility[list], and Answerability[list]. We collected 2-3 responses from different testees for each set. Participants received the same instructions as the models in order to compare them equally. More details can be found in Appendix~\ref{appendix:annotation}.

\subsection{Results}
\label{sec:results}
\begin{figure*}
    \centering
    \includegraphics[width=1\linewidth]{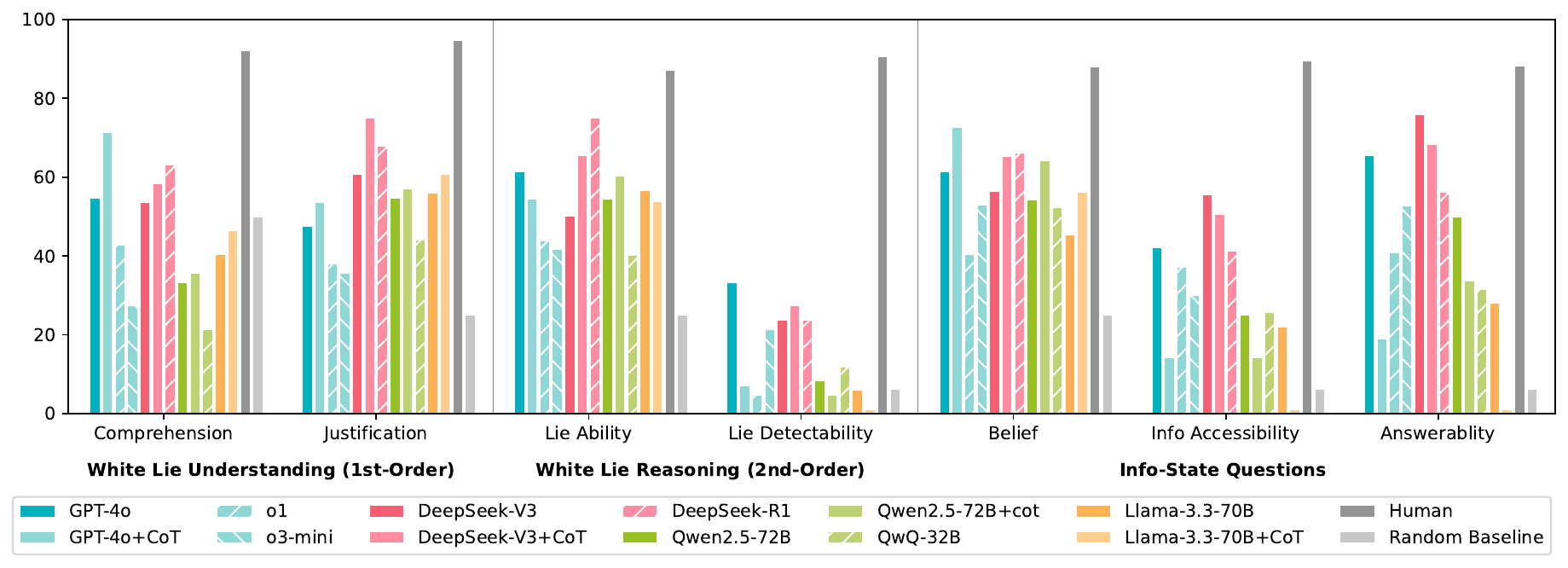}
    \caption{The answer accuracy of different LLM families on our benchmark: Comprehension[MCQs], Justification[MCQs], Lie Ability[MCQs], Lie Detectability[list], Belief[MCQs], Info Accessibility[list], and Answerability[list]}
    \label{fig:main-result}
\end{figure*}

Figure~\ref{fig:main-result} displays the full results of examined LLMs on TactfulToM. We categorize the results according to question types mentioned in Section~\ref{subsec:question-category} and use different colors to represent different models. Detailed scores are provided in Table~\ref{tab:model_performance} in Appendix~\ref{appendix:full_results}.

\paragraph{Overall Performance} GPT-4o and DeepSeek families consistently outperformed all other model families. DeepSeek models demonstrate a slight edge over GPT-4o on several tasks, including justification and Lie Ability questions. However, compared to humans who achieved an accuracy rate of over 85\% on all tasks, all current models still exhibit a substantial gap in our benchmark.

\paragraph{Vanilla vs. CoT Prompting vs. Reasoning Models} 
 CoT prompting shows inconsistent benefits across model families. GPT models show minimal improvements or even degrade performance with CoT prompting, particularly on lie detectability tasks. GPT reasoning models also unexpectedly underperformed their regular models. DeepSeek models exhibited a different pattern, with reasoning variants outperforming both vanilla models and CoT-prompted versions across most question categories. Llama and Qwen families demonstrated no consistent pattern in response to either CoT prompting or reasoning-specialized models. These findings suggest that current reasoning enhancement techniques provide inconsistent benefits for ToM reasoning involving white lies, indicating the need to improve performance in this domain.

\begin{figure}
    \centering
    \includegraphics[width=0.8\linewidth]{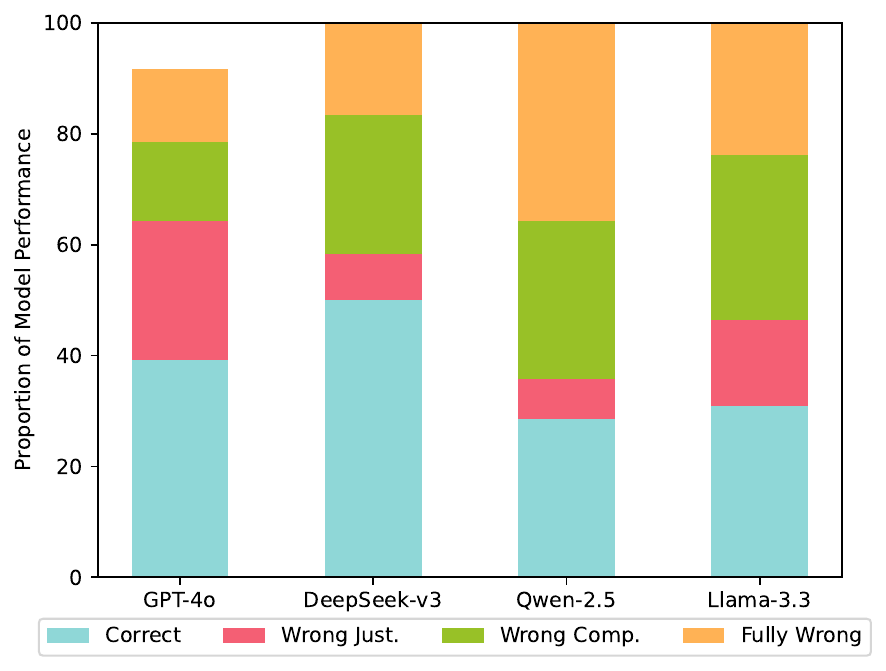}
    \caption{The proportion of model performance types in Justification questions.}
    \label{fig:error-comp-just}
\end{figure}

\paragraph{LLMs Struggle with True White Lie Understanding}
As described in Section~\ref{subsec:question-category}, true white lie understanding requires models to identify falsity while recognizing prosocial motivation. However, as shown in Figure~\ref{fig:error-comp-just}, model performance drops significantly on this combined task, with even the best models achieving $<50\%$ accuracy. This suggests that models may succeed on individual dimensions by chance or through pattern matching, without integrating the complementary aspects required for genuine understanding. DeepSeek-v3 performs best but remains far from human-level competence. Given that psychological research shows second-order ToM reasoning as a necessary condition for white lie understanding~\cite{happe1994advanced}, this result encourages further investigation into the second-order ToM reasoning capabilities of current LLMs.

\paragraph{LLMs Can Track Mental States But Fail to Apply Them in White Lie Contexts}
Our analysis reveals a performance gap between Info-State questions and White Lie Reasoning questions. While models track beliefs reasonably well, they struggle with questions requiring the application of these representations, particularly lie detectability where accuracy drops significantly. This pattern is consistent across all model families. This suggests two possibilities: either current LLMs possess mental state tracking abilities but cannot integrate these states to understand behavioral capabilities in white lie scenarios, or their apparent success in belief tracking may be superficial, lacking genuine second-order ToM reasoning needed to determine conditions for detecting deception.

\subsection{In-depth Analysis}
\label{sec:indepth_analysis}
\begin{figure*}
    \centering
    \includegraphics[width=1\linewidth]{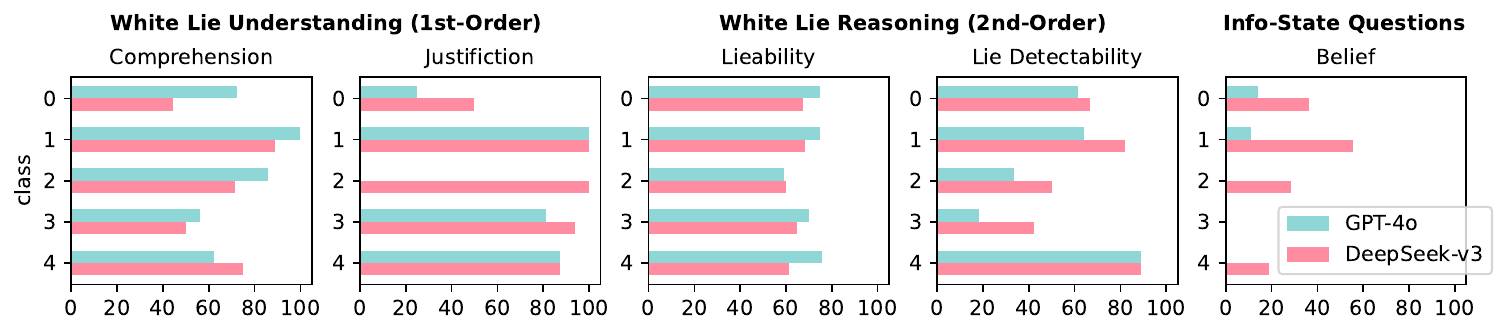}
    \caption{The performance of models across different classes. }
    \label{fig:classs}
\end{figure*}

\paragraph{Common Sense Falsehoods Are Easier for Models}
Our analysis reveals performance differences across different white lie classes as shown in Figure~\ref{fig:classs}. While Info-State questions show consistent performance, White Lie Understanding and Reasoning questions vary significantly. Models perform exceptionally well in Classes 1 and 2; this pattern suggests models use commonsense knowledge as a shortcut rather than engaging in genuine contextual reasoning. For Class 1 scenarios involving globally recognized falsehoods (e.g., ``Santa is real''), models can directly identify the statement as false without complex belief reasoning. Similarly, Class 2 scenarios featuring symbolic explanations of sensitive topics (e.g., death) are recognizable through common patterns in the data. In contrast, scenarios requiring situation-specific reasoning without obvious common sense cues pose significantly greater challenges, highlighting that models still largely rely on statistical regularities rather than sophisticated ToM capabilities when navigating white lie understanding.

\paragraph{Surface-Level Detection vs. Motivation Understanding}
Table \ref{tab:model-type-performance} shows a clear drop in model performance across three falsifiability levels (described in Section~\ref{subsec:structuring-white-lies}). DeepSeek-v3's comprehension, for instance, falls from Level-1 (79.41\%) to Level-3 (34.78\%). This reveals a critical insight: models excel at detecting lies through explicit contradictions but struggle to infer deception directly from motivations. A manual examination of DeepSeek's reasoning reveals the model primarily identifies lie detectability by checking which characters have access to what information. This pattern explains DeepSeek-v3's counterintuitive improvement in lie detectability for Level-3 (39.13\%) compared to Level-1 (26.47\%): without explicitly stated truths, the model faces less confusion about characters' information access but fails to recognize that genuine white lie detection requires understanding protective intentions, not merely contradiction recognition.

\begin{table}[!t]
  \centering
  \scriptsize
  \renewcommand{\arraystretch}{1.2}
  \begin{tabular}{p{0.9cm} p{0.5cm} p{0.6cm} p{0.6cm} p{0.6cm} p{0.8cm} p{0.8cm}}  
    \toprule
    \textbf{Model} & \textbf{Level} & \textbf{Comp} & \textbf{Justi} &  \textbf{B-2} & \textbf{LieAb} & \textbf{LieDe}\\
    \midrule
    \multirow{3}{*}{\textbf{GPT-4o}} 
    & L-1 & 79.41 & 55.88 & 69.02 & 51.06 & 8.82\\
    & L-2 & 70.37 & 59.26 & 73.58 & 63.08 & 7.41\\
    & L-3 & 60.87 & 43.48 & 71.50 & 47.06 & 4.35\\
    \midrule
    \multirow{3}{*}{\textbf{DeepSeek}} 
    & L-1 & 79.41 & 88.24 & 62.63 & 65.96 & 26.47\\
    & L-2 & 51.85 & 66.67 & 69.81 & 64.62 & 18.52\\
    & L-3 & 34.78 & 65.22 & 55.56 & 58.82 & 39.13\\
    \bottomrule
  \end{tabular}
  \caption{Performance (\%) of GPT-4o and DeepSeek-v3 across levels. Abbreviations: Comp=Comprehension, Justi=Justification, B-2=2nd-order Belief, LieAb=Lie Ability, LieDe=Lie Detectability.}
  \label{tab:model-type-performance}
\end{table}

\paragraph{Models Struggle with Genuine Motivation Understanding Without Guidance}
To assess models' true comprehension of white lie motivations, we examined the Justification question's free-form responses where no options provide hints. As shown in Table \ref{tab:question_format}, models' performance drops significantly from MCQs to free-form responses. DeepSeek's falls from 75\% to approximately 30\% across different metrics. This gap suggests multiple-choice accuracy is inflated by provided options, as models struggle to independently infer the prosocial intentions behind white lies. Even the best-performing models fail to identify emotional protection motivations in most free-form responses, highlighting significant limitations in their unprompted emotional reasoning.

\begin{table}[!t]
  \tabcolsep=3.5pt
  \centering
  \scriptsize
  \renewcommand{\arraystretch}{1.05}
      \begin{tabularx}{0.48\textwidth}{p{0.9cm}>{\centering\arraybackslash}X>{\centering\arraybackslash}X>
      {\centering\arraybackslash}X>{\centering\arraybackslash}X}
        \toprule
        \multirow{2}{*}{\textbf{Model}} & \multirow{2}{*}{\textbf{MCQs}} & \multicolumn{3}{c}{\textbf{FreeForm}}\\
        \cmidrule{3-5}
        & & \textbf{Cos. Sim.} & \textbf{Token-F1} & \textbf{LLM-Judge}\\
        \midrule
        \textbf{GPT-4o} & 53.57 & 22.62 & 27.38  & 16.67\\
        \textbf{DeepSeek} & 75.00 & 29.76 & 35.71 & 26.19\\
        \textbf{Qwen} & 57.14 & 19.05 & 9.52 & 25.00\\
        \textbf{Llama} & 46.43 & 20.24 & 10.71 & 23.81\\
        \bottomrule
      \end{tabularx}
  \caption{The accuracy of the model's CoT performance in Justification tasks under different task formats and evaluation methods.}
  \label{tab:question_format}
\end{table}

\section{Related Work}
\paragraph{ToM in Psychology} 
ToM has been categorized into seven abilities by the Abilities in Theory of Mind Space framework~\cite{beaudoin2020systematic}: Intentions, Desires, Emotions, Knowledge, Percepts, Beliefs, and Mentalistic Understanding of Non-literal Communication. Non-literal communication understanding enables the interpretation of non-literal language such as irony, sarcasm, and white lies, where the intended meaning diverges from the literal. This requires second-order ToM ability, which is typically assessed through false-belief tasks and nested belief attribution~\citep{wimmer1983beliefs, quesque2020theory} in psychology experiments. As \citet{beaudoin2020systematic} demonstrated, accurately interpreting such expressions depends on the listener's capacity to infer communicative intent and consider the speaker's emotional motivations. These forms of pragmatic inference are especially relevant in white lies, where the goal may be to avoid harm or maintain relationships~\citep{erat2012white}. This reflects a broader understanding of ToM as a key mechanism for navigating complex social communication, supported by evidence from developmental, clinical, and neurocognitive studies~\citep{baron1985does, langley2022theory}.

\paragraph{ToM in LLMs}
Most existing ToM evaluations of LLMs focus on false-belief tests, such as the benchmarks ToMi~\cite{nematzadeh-etal-2018-evaluating}, ToM-QA~\cite{le-etal-2019-revisiting}, and FANToM~\cite{kim-etal-2023-fantom}, primarily testing whether models can track belief states when objects are moved or information changes. 
Some recent works go beyond false belief tests by incorporating broader mental states testing~\cite{chen-etal-2024-tombench, shinoda2025tomato} and explore ToM in applied social scenarios~\cite{chan-etal-2024-negotiationtom, bara-etal-2021-mindcraft, shapira-etal-2023-well}, or improving evaluation methods~\cite{10.5555/3666122.3666717}.
However, white lies remain largely understudied. While ToMBench~\cite{chen-etal-2024-tombench} included white lie tests, it offers only 20 non-conversational samples, too limited for comprehensive evaluation. This limited understanding of white lie capabilities poses risks as LLMs are increasingly deployed in emotional support and caregiving applications where such skills are essential. 

\section{Conclusion} 
We present TactfulToM, an English ToM benchmark designed to evaluate LLMs' understanding of white lies through complex social scenarios. Our comprehensive evaluation reveals that even state-of-the-art LLMs underperform compared to humans in white lie understanding and reasoning, particularly in understanding the emotional motivation behind it.
This performance gap raises ethical questions about LLMs' development: should LLMs understand white lies merely to interpret human behavior, or also to potentially generate them? The dilemma lies in choosing between strict truthfulness and social grace that might involve benign deception. TactfulToM provides a foundation for improving LLMs' social reasoning of white lie understanding, but we must carefully consider whether aligning LLMs completely with human social behaviors, including prosocially-motivated deception, is truly desirable for human-AI interaction.
\section*{Limitations}
The main limitations of this paper are:
\paragraph{Limited to White Lies}
This dataset is primarily focused on white lie scenarios in order to analyze LLMs’ ToM capabilities in such contexts. We do not extensively explore LLMs’ other second-order ToM abilities; however, we hope that the methodology proposed in this paper can provide insights for future researchers seeking to construct related datasets.

\paragraph{Lack of Prior Impression}
In real-life situations, people typically possess prior knowledge and impressions of others. In our dataset, we deliberately constrained the scenarios such that the white lie triplets are not previously known to any of the involved roles, with the exception of the liar who initiates the deception. While this design choice helps isolate the ToM reasoning process, it does not fully capture the complexity of real-world social interactions. We consider incorporating this aspect of human cognition in our future work.

\paragraph{Limited Culture and Language}
Our benchmark includes only English-language data. However, in some other languages and cultures, communication tends to be more indirect, which may lead to different patterns of ToM reasoning in white lie scenarios.

\section*{Societal and Ethical Considerations}
We acknowledge that our focus on white lies and Theory of Mind may raise concerns about anthropomorphizing AI systems. However, our research does not advocate for developing AI systems capable of telling white lies. Rather, we aim to systematically evaluate LLMs' social reasoning capabilities within specific informational contexts. Our results demonstrate that current models fall significantly short of human-like understanding in these scenarios, primarily relying on pattern matching rather than genuine understanding of mental states or intentions. We recognize the ethical complexities surrounding deception, even when prosocially motivated, and the particular sensitivity of developing AI systems with capabilities that could involve any form of misrepresentation.

All annotators participating in our data collection and validation were recruited through Prolific. We established fair compensation standards based on estimated task duration, ensuring payment rates above minimum wage requirements. We maintained transparent communication channels with annotators, promptly addressing questions and incorporating feedback to improve task instructions. All annotator data was anonymized, with only minimal identifiers stored securely and not included in the released dataset.
We were careful to design our task instructions clearly, providing sufficient context without biasing responses. Annotators were informed about the academic research nature of the task and how their contributions would be used. When selecting annotators, we sought diversity across demographic factors to minimize potential biases in our data collection process, though we acknowledge that online recruitment platforms have inherent demographic limitations.

Our dataset is intended for research purposes only. While we have taken measures to ensure the conversations do not contain offensive content, research using generative models always carries a risk of unexpected outputs, particularly in free-form reasoning contexts. We encourage responsible use of our benchmark and dataset for advancing understanding of social reasoning in AI systems while remaining mindful of potential misapplications.

\section*{Acknowledgments}
This work was supported by JST FOREST Grant Number JPMJFR232R and JST K Program Grant Number JPMJKP24C2. Thanks to EPFL College of Humanities for providing financial support for Yiwei Liu to attend the conference and present this work. We would like to thank the anonymous reviewers for their valuable feedback and comments that helped improve this paper, also the participants who contributed to the annotation.

\bibliography{anthology,custom}

\appendix

\section{TactfulToM Construction}

\subsection{Dataset Distribution}
\label{appendix:dataset_distribution}
\begin{figure}[!tb]
    \centering
    \includegraphics[width=1\linewidth]{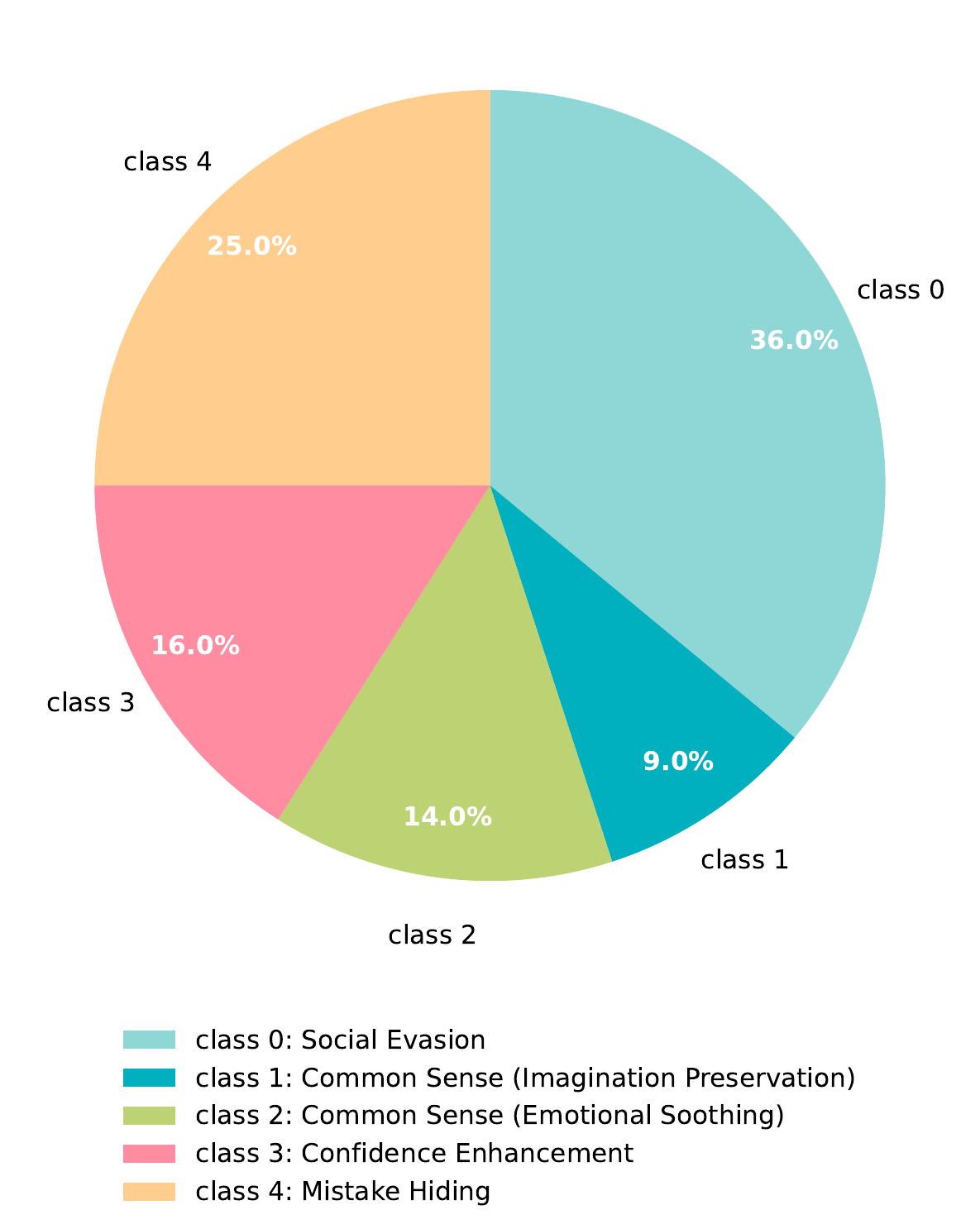}
    \caption{Proportion of different classes in TacfulToM.}
    \label{fig:proportion_classes}
\end{figure}

The proportion distribution of different classes within the TacfulToM dataset is shown in Figure \ref{fig:proportion_classes}.

\subsection{Prompt Templates for Conversation Generation}
Prompt templates for generating the conversation are listed below as Step\_1, Step\_2, Step\_3, and Step\_4. 
\label{appendix:conversation prompt template}
\begin{tcolorbox}[colback=gray!5!white, colframe=black, title=Step\_1, fontupper=\small]
In this conversation, four characters are participating: {{A: the liar name}}, {{B: the target name}}, {{C: the accomplice name}} and {{D: the observer name}}.

Start the conversation with all participants discussing {{Topic for the scenario}}.

They are {{relationship descriptor}}, {{A: the liar name}} and {{C: the accomplice name}} are parents, {{B: the target name}} is kid, and {{D: the observer name}} is aunt/uncle.
\label{Step-1}
\end{tcolorbox}

\begin{tcolorbox}[colback=gray!5!white, colframe=black, title=Step\_2, fontupper=\small]
Now, {{B: the target name}} leaves the conversation because of the reason: {{leave reason B}}, while {{D: the observer name}} leaves the conversation because of the reason: {{leave reason D\_1}}.

With {{B: the target name}} and {{D: the observer name}} absent, {{A: the liar name}} and {{C: the accomplice name}} continue their conversation, shifting the topic to '{{real\_reason\_c}}'. Ensure the dialogue flows naturally without directly repeating these exact words.

Remember to indicate that {{B: the target name}} and {{D: the observer name}} are leaving the conversation.
\end{tcolorbox}

\begin{tcolorbox}[colback=gray!5!white, colframe=black, title=Step\_3, fontupper=\small]
Now, only {{D: the observer name}} returns and rejoins the conversation after previously leaving. Note that {{B: the target name}} has not returned to the conversation yet.

Then the conversation continues and naturally shifts to {{situation\_topic}}. Do NOT catch up or recap details from the previous conversation.

And then '{{truth\_c}}' is naturally revealed in the conversation.

After engaging briefly, {{D: the observer name}} leaves the conversation again because of the reason: {{leave reason D\_2}}.

IMPORTANT:
\begin{itemize}
  \item Clearly indicate through dialogue only (e.g., "I'm back," or "I have to leave again now.") that {{D: the observer name}} first returns and later departs again.
  \item Do NOT use third-person narrative or stage directions.
  \item Avoid mentioning details from previous conversations.
\end{itemize}
\end{tcolorbox}

\begin{tcolorbox}[colback=gray!5!white, colframe=black, title=Step\_4, fontupper=\small]
Now {{B: the target name}} returns to the conversation after leaving the conversation.

First have {{B: the target name}} explicitly indicate the return through dialogue. Do NOT catch up or recap details from the previous conversation.

And then {{situation}} naturally unfolds. Make sure the dialogue flows naturally without directly repeating these exact words.

In response, {{A: the liar name}} and {{C: the accomplice name}} tells {{B: the target name}} that '{{the lie}}'.

IMPORTANT:
\begin{itemize}
  \item Do NOT use third-person narrative or stage directions.
  \item Avoid mentioning details from previous conversations.
\end{itemize}
\end{tcolorbox}

\subsection{Leaving Reason List}
\label{appendix:leaving_reason}

\noindent\rule{\columnwidth}{0.5pt} 
\vspace{0em}
bathroom break \\
coffee break \\
forgot something important \\
forgot to print some documents \\
forgot to receive a package \\
forgot to return a package \\
forgot to run errands \\
forgot to submit documents \\
have a meeting starting soon that I need to prepare for \\
have a previous engagement that I need to attend to quickly \\
have a work-related emergency that requires my immediate attention \\
have an unexpected visitor at my door \\
have errands to run \\
have to attend to someone who just walked in \\
have to check on something \\
have to go to the restroom \\
have to pick up a prescription \\
have to pick up dry cleaning \\
have to print or scan documents \\
have to receive a delivery \\
have to recharge laptop \\
have to return a borrowed item \\
have to take care of a family matter \\
have to take care of an unexpected task \\
have unexpected visitor \\
his/her pet needs attention \\
his/her family is calling \\
incoming delivery \\
must respond to a phone call \\
need to check on a friend or family member who needs assistance \\
need to finish a task that’s time-sensitive \\
need to get a phone call \\
need to get some coffee \\
need to go to the toilet \\
need to grab a snack or a drink \\
need to have a quick chat with someone else \\
need to make a phone call \\
need to make a quick trip to the drug store \\
need to make a quick trip to the grocery store \\
need to pick up a package \\
need to receive a parcel \\
need to recharge cellphone \\
need to register for an event \\
need to schedule a haircut or salon appointment \\
need to schedule another appointment \\
need to step out for a moment \\
need to submit some papers \\
need to take care of some paperwork or documents \\
need to take care of some personal matters \\
need to take care of something urgent \\
need to troubleshoot something \\
parking meter expiring \\
remembered something that needs to be taken care of \\
remembered to receive a package \\
remembered to submit some papers \\
remembered to take care of some paperwork or documents \\
remembered to take care of some personal matters \\
remembered to take care of something urgent \\
want to go grab a drink \\
want to go grab a coffee \\
want to go take some fresh air \\
want to go to the bathroom \\
need to move my car \\
have to take an urgent call from my boss \\
need to check my emails quickly \\
have to respond to an important message \\
need to restart my computer \\
have to take a quick medication \\
need to handle a minor household emergency \\
have to refill my water bottle \\
have to feed my pet \\
have to water my plants \\
have to take a brief walk to clear my mind \\
need to step outside briefly to meet someone \\
have to adjust the thermostat \\
need to quickly tidy up my workspace \\
need to quickly verify something important \\
have to quickly arrange something for a later meeting \\
need to briefly step out to confirm travel arrangements \\
have to take care of an urgent email \\
have to briefly assist a coworker \\
need to briefly leave to verify appointment details \\
have to briefly tend to something outside \\
have to quickly reschedule an upcoming meeting \\
need to briefly attend to my child \\
need to grab a quick snack \\
need to quickly stretch my legs \\
have to briefly troubleshoot my internet connection \\
have to briefly step out for privacy reasons \\
need to quickly tidy the room before another meeting \\
have to quickly update someone about my status \\
need to briefly review notes or materials \\
have to briefly leave to answer an urgent text \\
have to briefly assist someone else in the household \\
have to quickly plug in my device to charge \\
need to briefly leave to retrieve an important item \\
have to step away briefly to close a window or door \\
need to quickly ensure I turned off an appliance \\
have to briefly step away for a personal reason \\
need to briefly step away due to allergy or health symptoms \\
\vspace{0em}
\noindent\rule{\columnwidth}{0.5pt}
\captionof{table}{Predefined reasons for characters leaving the conversation.}
\label{tab:leaving_reasons_flow}

\subsection{Clarification on the usage of gpt-4o for dataset generation}
\label{appendix:clarification}
The generation pipeline relies on GPT-4o, while our pre-experimental results show it struggles with detecting white lies. This raises concerns about the appropriateness of using LLMs for generating conversational data involving white lies. However, this is precisely why we designed our white lie triplet and 4-step generation pipeline—to avoid having the model understand what constitutes a white lie and then generate conversations based on the model's potentially flawed understanding of white lies. Our white lie triplet decomposes white lies into three core elements (the lie, the truth, and the real reason). The 4-step generation loop uses one element at each step, requiring the model to understand only individual components rather than sophisticated white lie reasoning. This method ensures the model builds conversations without being instructed to ``write a white lie.''

\subsection{Question Generation Templates}
\label{appendix:question_template}
The question generation templates we used are provided below as examples in Table~\ref{tab:question_template}.

\begin{table}[!hbt]
    \centering
    \small
    \begin{tabular}{p{0.25\linewidth}p{0.65\linewidth}}
    \toprule
    \textbf{Question Type} & \textbf{Template} \\
    \midrule
    \textbf{Comprehension} & {``Was what \{liar\} told \{target\}, \{lie\}', true?''} \\ \midrule \textbf{Justification} & {``Why did \{liar\} and \{accomplice\} tell \{target\} \{lie\}'?''} \\
    \midrule
    \textbf{Lie Ability} & {``Please choose the correct answer from the options below:''\newline
    Correct: ``\{liar\} tell \{target\} that \{lie\}' because \{liar\} believes that \{target\} is unaware that \{real reason\}.''\newline Wrong 1: \{liar\} tell \{target\} that \{lie\}' because \{liar\} believes that \{target\} believes \{real reason\}.''\newline
    Wrong 2: ``\{liar\} tell \{target\} that \{lie\}' because \{liar\} is unaware that \{target\} is unaware that \{real reason\}.''\newline Wrong 3: \{liar\} tell \{target\} that \{lie\}' because \{liar\} is unaware that \{target\} believes \{real reason\}.''} \\
    \midrule
    \textbf{Lie Detectability} & {``Was what \{liar\} told \{target\}, \{lie\}', true?''\newline List all the characters who know the precise correct answer to this question.''} \\ \midrule \textbf{Fact} & {``Who discussed \{real reason\}'?''\newline
    ``Who discussed \{truth\}'?''} \\ \midrule \textbf{Belief} & {``What does {character} believe about \{truth question\} or \{real reason question\}?''\newline ``What does \{liar\} believe about \{target\}'s belief regarding \{truth question\} or \{real reason question\}'?''} \\
    \midrule
    \textbf{Info Accessibility (List)} & {``List all the characters who know this information: \{real reason\}''} \\
    \midrule
    \textbf{Info Accessibility (Binary)} & {``Does \{character\} know this information: \{real reason\}?''} \\
    \midrule
    \textbf{Answerability (List)} & {``Who discussed \{real reason\}'?''\newline ``List all the characters who know the precise correct answer to this question.''} \\ \midrule \textbf{Answerability (Binary)} & {``Who discussed \{real reason\}'?''\newline
    f``Does \{character\} know the precise correct answer to this question?''} \\
    \bottomrule
    \end{tabular}
    \caption{Question generation templates Examples for different question types in TactfulToM.}
    \label{tab:question_template}
\end{table}

\subsection{Wrong Option Design}
\label{appendix:wrong-options}

\paragraph{Belief Statement Options}
For second-order belief statements, we formalized four logically distinct cases using belief operators:

\begin{itemize}
    \item $Bel_Z(\varphi)$: Person Z believes proposition $\varphi$
    \item $\neg Bel_Z(\varphi)$: Person Z is unaware of (or does not believe) $\varphi$
\end{itemize}

A second-order belief statement takes the form $Bel_X(\cdot)$, where the inner argument concerns Y's epistemic state about proposition p:
\begin{equation*}
\small
\begin{aligned}
    & Bel_X(Bel_Y(p)) && \text{(X thinks Y thinks p)} \\
    & Bel_X(\neg Bel_Y(p)) && \text{(X thinks Y is unaware of p)} \\
    & \neg Bel_X(Bel_Y(p)) && \text{(X is unaware that Y thinks p)} \\
    & \neg Bel_X(\neg Bel_Y(p)) && \text{(X is unaware that Y is unaware of p)}
\end{aligned}
\end{equation*}

When ``X thinks Y thinks p'' is supported by the dialogue, we use $Bel_X(Bel_Y(p))$ as the correct answer. The remaining expressions serve as distractors representing three error types:
\begin{itemize}
    \item Wrong attribution of Y's first-order belief
    \item Wrong attribution of X's meta-belief
    \item Simultaneous error in both belief layers
\end{itemize}

\paragraph{Justification Options}
\label{appendix:Justification-options}

For justification questions, we employed few-shot prompting with the following criteria:

\begin{itemize}
    \item Correct answer must reflect the genuine prosocial motivation (e.g., sparing feelings, maintaining harmony)
    \item Wrong options:
    \begin{itemize}
        \item Mutually exclusive
        \item Plausible alternative explanations
        \item Consistent with the dialogue context
    \end{itemize}
\end{itemize}

Example:
\begin{quote}
Correct: ``Because Jamey wants to politely decline without making Pearl feel bad about choosing an expensive restaurant.''

Wrong options:
\begin{itemize}
    \item ``Because Jamey actually has to work this weekend.''
    \item ``Because Jamey dislikes Pearl and doesn't want to spend time with him.''
    \item ``Because Jamey already has dinner plans with someone else.''
\end{itemize}
\end{quote}

\subsection{Quality control of generation}
\label{appendix:generation_quality}
Generation filter: Each dialogue has four parts. Because later parts are conditioned on earlier ones, we observed quality decrease in later steps. To maintain high standards, conversation parts (steps 2-4) typically required 3-4 iterations or more to achieve natural, realistic dialogue flow. This iterative approach ensured that only high-quality conversations reached the annotation phase.

\subsection{An Example from TactfulToM}
\label{appendix:datasample}
We provided a full conversation sample from TactfulToM below for reference:

\noindent\rule{\columnwidth}{0.5pt} 
\vspace{0em}
\textbf{Pearl}: So, I was thinking about food and I'm curious, what's everyone's favorite cuisine? I absolutely love Italian, especially a good risotto. It just feels like a warm hug in a bowl!

\textbf{Jamey}: Oh, Italian is great! But for me, it's definitely Thai food. I love the bold flavors and the perfect balance of sweet, sour, and spicy. Pad Thai is my absolute favorite.

\textbf{Donyae}: Oh, all great choices! I'm all in for Mexican food. Tacos, enchiladas, and don't even get me started on the guacamole. It's just so satisfying.

\textbf{Pearl}: Yum, I could eat some good guac every day. Has anyone tried making it from scratch? It's surprisingly easy and so much better than store-bought.

\textbf{Jamey}: I have! I actually make a mean guacamole—and you're right, Pearl, fresh ingredients make all the difference. It's a hit whenever I have friends over.

\textbf{Marithza}: I bet! Guacamole's a crowd pleaser for sure. On another note, I tried making homemade sushi once. Let's say, rolling those little vinegared rice gems isn't as easy as it looks!

\textbf{Donyae}: I can imagine! But doing it yourself must be fun. Maybe we should have a cooking night sometime, where everyone brings a dish from their favorite cuisine.

\textbf{Pearl}: That's a fantastic idea, Donyae! We could even make it like a friendly mini competition. Each dish gets a few points for taste and presentation. What do you guys think?

\textbf{Jamey}: I'm totally in! It'll be a fun way to try different cuisines and maybe even learn a few new recipes from each other.

\textbf{Donyae}: Awesome! I just remembered, I need to submit some papers before the end of the day. I'll catch you guys later, though. This cooking night sounds like such a blast—keep me posted!

\textbf{Pearl}: Oh, I need to make a quick phone call. Let's definitely plan this out soon! Bye for now, everyone.

\textbf{Marithza}: Bye, Donyae and Pearl! We'll keep you in the loop. So, Jamey, what kind of place were you thinking about for our cooking night? We could do it at someone's house or rent out a small venue.

\textbf{Jamey}: Honestly, I love the idea of just doing it at someone's house. It keeps things cozy and informal. Kind of like why I prefer casual dining places over the upscale spots. There's something special about a laid-back atmosphere.

\textbf{Marithza}: I know what you mean. Sometimes those upscale places can feel a bit too… stiff, maybe? I guess it's nice to just unwind and not worry about which fork to use—chat over some good food without all the fancy stuff.

\textbf{Jamey}: Exactly! Plus, it can be pretty expensive going out to those places too often. I'd much rather hit up a cozy little diner or a local joint where you can really taste the love put into the cooking.

\textbf{Marithza}: For sure! And there's something so genuine about finding that hole-in-the-wall spot with the best food. It kind of feels like discovering a secret everyone should know about.

\textbf{Donyae}: Hey everyone, I'm back! I finally submitted all the papers. What did I miss?

\textbf{Marithza}: Welcome back, Donyae! We were just talking about where to hold our cooking night. Jamey and I thought doing it at someone's house might be fun and cozy.

\textbf{Donyae}: That sounds perfect! I think the home setting adds a nice personal touch. Plus, it's easier to relax and enjoy each other's company.

\textbf{Jamey}: Absolutely, home-cooked vibes really can't be beat. Do you have a dish in mind for when we do this?

\textbf{Donyae}: I'm thinking something classic like enchiladas. They are always a hit and not too complicated. By the way, I have to leave again now—I have a meeting starting soon that I need to prepare for.

\textbf{Marithza}: No worries, Donyae. Thanks for popping back in! Let us know how the meeting goes.

\textbf{Jamey}: Yeah, hope all goes well, Donyae! We'll continue planning and keep you updated.

\textbf{Pearl}: Hey everyone, I'm back from my call! What did I miss in the world of culinary planning?

\textbf{Jamey}: Welcome back, Pearl! We were chatting about turning our houses into five-star restaurants for our cooking night.

\textbf{Pearl}: Sounds exciting! But speaking of five-star, there's this new Italian restaurant that just opened downtown. Would you all be interested in checking it out this weekend?

\textbf{Jamey}: That sounds amazing, Pearl, but unfortunately, I have to work this weekend, so I can't make it. Maybe next time?

\textbf{Pearl}: Oh no, that's a bummer! We'll definitely catch up soon then. Maybe for that cooking night we talked about—we can even bring some Italian-inspired dishes to you instead.

\textbf{Marithza}: I'm up for the restaurant visit if it's still on. I've been dying to try their truffle pasta from what I've heard.

\textbf{Pearl}: Awesome, Marithza! Let's make it a date then. We'll let Donyae know and hopefully, she can join us too.

\textbf{Jamey}: You guys enjoy it! Be sure to save me a slice of that truffle pasta, at least in spirit.

\textbf{Marithza}: We will! And we'll definitely share all the delicious details with you. Catch up soon, Jamey!

\textbf{Pearl}: For sure, Jamey. Good luck with work, hope the weekend goes smoothly!

\vspace{0em}
\noindent\rule{\columnwidth}{0.5pt} 

\section{Experiemnts}
\subsection{Annotation details}
\label{appendix:annotation}

\paragraph{Annotator Recruitment}
Our participants for coherence validation were recruited on the Prolific platform\footnote{\url{https://www.prolific.com/}}. They met the following criteria: English as their first and primary language, fluency in English, and completion of an undergraduate degree (BA/BSc/other). Additionally, participants had an approval rate between 95–100\%. We posted a qualification screening test with 79 initial candidates. The screening test required participants to evaluate conversation coherence for three dialogues, two of which were intentionally scrambled for a few sentences. Applicants were required not only to identify incoherent conversations but also to provide explanations for why they considered the conversations incoherent. This screening process filtered the pool down to 21 reliable annotators to ensure the quality of human annotation, who served as participants for the coherence validation. We paid participants 9 pounds per hour, which is considered appropriate according to Prolific's standards. This payment level ensures fair compensation for participants, who were predominantly located in English-speaking regions, thereby encouraging high-quality annotations.

For human performance annotation, the participants consisted of annotators (crowdworkers) recruited via Prolific and graduate students. We recruited a total of 9 qualified annotators through the Prolific platform. We additionally included 3 graduate students to diversify the participant pool and enhance annotation quality. The inclusion of graduate students allowed us to validate the reliability of crowdsourced annotations and maintain consistency across all raters.

\paragraph{Annotation Process for Coherence}
Each conversation is reviewed for coherence, safety, and white lie authenticity using binary assessment with written explanations for flags.
Each conversation was reviewed by three annotators, and we required a 2/3 majority approval rate to retain conversations in the final dataset. While 7 out of 100 dialogues received individual flags, none reached majority disagreement.

\paragraph{Human Performance Evaluation}
For conversation-wise sampling, we randomly sampled 15 out of 100 conversation sets using stratified sampling to maintain representation across all five classes and three difficulty levels (described in Section~\ref{subsec:structuring-white-lies}). The number of sampled sets (15/100) was determined by referencing the human evaluation ratio in similar ToM benchmarks - for example, FANToM~\cite{kim-etal-2023-fantom} evaluated 32 out of 256 sets (12.5\%), while our ratio (15\%) provides comparable coverage.  For each sampled conversation set, we collected 2-3 human responses to ensure reliability.

For question type-wise collection, human data was collected for only one format of each question type (e.g., only MCQ format, not freeform). This design choice reflects our human evaluation methodology, where participants received one conversation along with all associated questions for that conversation, making it redundant to ask the same conceptual question in multiple formats. Our model-human comparisons are therefore conducted only on equivalent question formats.

\subsection{Model Performance on All Tasks}
\label{appendix:full_results}
Detailed scores of the model performance on all tasks are provided in Table~\ref{tab:model_performance}.

\begin{table*}[!htb]
  \tabcolsep=3.5pt
  \centering
  \tiny
  \renewcommand{\arraystretch}{1.05}
      \begin{tabularx}{\textwidth}{p{1.2cm}>{\centering\arraybackslash}X>{\centering\arraybackslash}X>
      {\centering\arraybackslash}X>
      {\centering\arraybackslash}X>{\centering\arraybackslash}X>{\centering\arraybackslash}X>{\centering\arraybackslash}X>
      {\centering\arraybackslash}X>
      {\centering\arraybackslash}X>
      {\centering\arraybackslash}X>
      {\centering\arraybackslash}X>{\centering\arraybackslash}X>
      {\centering\arraybackslash}X>{\centering\arraybackslash}X>{\centering\arraybackslash}X>{\centering\arraybackslash}X>
      {\centering\arraybackslash}X>
      {\centering\arraybackslash}X>
      {\centering\arraybackslash}X}
        \toprule
          \multirow{2}{2em}{\textbf{Model}} & \multirow{2}{2em}{\textbf{Class}}& \multicolumn{2}{c}{\textbf{Comp}} & \multicolumn{2}{c}{\textbf{Justification}}& \textbf{LieAb.} & \multicolumn{2}{c}{\textbf{Lie Detectability}} &  \multicolumn{2}{c}{\textbf{Belief}} & \multicolumn{2}{c}{\textbf{Info Accessibility}} & \multicolumn{2}{c}{\textbf{Answerablity}} & \multicolumn{2}{c}{\textbf{FactReason}}  & \multicolumn{2}{c}{\textbf{FactTruth}}  \\
            & &MCQs& Free & MCQs& Free& MCQs & List&Binary  &  MCQs& Free & List&Binary & List&Binary&MCQs& Free  & MCQs& Free  \\
        \midrule
Human&  &92.11 & - & 94.74 & - & 87.23 & 90.65 & - & 88.16 & - & 89.50 & - & 88.34 & - & - & - & - & - \\
\midrule
GPT-4o&\multirow{12}{*}{0} &41.67 & 19.44 &  11.11 & 22.22& 60.26 & 36.11 & 51.85& 62.36 & 41.95 & 34.48 & 75.97 & 62.07 & 62.07  & 100.0 & 63.89 & 86.36 & 63.64  \\
\textit{+CoT}& &72.22 & 25.0 &  25.0 & 16.67& 61.54 & 13.89 & 62.04& 74.86 & 49.71 & 20.69 & 88.96 & 15.52 & 15.52  & 100.0 & 58.33 & 63.64 & 68.18  \\
o1& &30.56 & 75.0 &  5.56 & 33.33& 34.62 & 8.33 & 36.11& 31.18 & 32.76 & 36.21 & 48.7 & 39.66 & 39.66  & 50.0 & 41.67 & 31.82 & 27.27  \\
o3-mini& &16.67 & 11.11 &  2.78 & 5.56& 38.46 & 30.56 & 71.3& 51.87 & 36.49 & 27.59 & 68.83 & 51.72 & 51.72  & 97.22 & 55.56 & 59.09 & 50.0  \\
DeepSeek-V3& &36.11 & 13.89 &  25.0 & 19.44& 51.28 & 27.78 & 57.41& 57.9 & 40.23 & 41.38 & 65.58 & 82.76 & 82.76  & 100.0 & 63.89 & 68.18 & 54.55  \\
\textit{+CoT}& &44.44 & 38.89 &  50.0 & 27.78& 66.67 & 36.11 & 65.74& 67.53 & 54.17 & 60.34 & 88.31 & 75.86 & 75.86  & 94.44 & 69.44 & 50.0 & 72.73  \\
DeepSeek-R1& &52.78 & 25.0 &  44.44 & 19.44& 73.08 & 27.78 & 70.37& 66.95 & 49.43 & 36.21 & 87.66 & 56.9 & 56.9  & 100.0 & 58.33 & 77.27 & 50.0  \\
Qwen2.5& &8.33 & 27.78 &  8.33 & 25.0& 55.13 & 5.56 & 73.15& 47.7 & 36.35 & 24.14 & 64.29 & 44.83 & 44.83  & 100.0 & 58.33 & 54.55 & 54.55  \\
\textit{+CoT}& &16.67 & 33.33 &  11.11 & 11.11& 62.82 & 5.56 & 65.74& 63.36 & 39.37 & 18.97 & 61.69 & 24.14 & 24.14  & 97.22 & 83.33 & 50.0 & 45.45  \\
QwQ& &2.78 & 61.11 &  8.33 & 33.33& 14.1 & 5.56 & 27.78& 35.92 & 28.74 & 12.07 & 44.16 & 20.69 & 20.69  & 47.22 & 25.0 & 27.27 & 13.64  \\
Llama-3.3& &19.44 & 19.44 &  8.33 & 5.56& 60.26 & 11.11 & 75.0& 45.55 & 48.28 & 18.97 & 74.03 & 39.66 & 39.66  & 100.0 & 55.56 & 59.09 & 63.64  \\
\textit{+CoT}& &41.67 & 38.89 &  16.67 & 16.67& 48.72 & 0.0 & 63.89& 59.63 & 44.97 & 0.0 & 76.62 & 0.0 & 0.0  & 100.0 & 58.33 & 77.27 & 59.09  \\
\midrule
GPT-4o&\multirow{12}{*}{1} &100.0 & 55.56 &  100.0 & 11.11& 82.14 & 66.67 & 55.56& 67.86 & 36.31 & 50.0 & 75.0 & 78.57 & 78.57  & 100.0 & 77.78 & 80.0 & 60.0  \\
\textit{+CoT}& &100.0 & 77.78 &  100.0 & 0.0& 64.29 & 11.11 & 62.96& 75.0 & 40.48 & 7.14 & 78.57 & 28.57 & 28.57  & 88.89 & 77.78 & 60.0 & 60.0  \\
o1& &100.0 & 77.78 &  100.0 & 33.33& 92.86 & 0.0 & 100.0& 67.26 & 39.29 & 42.86 & 71.43 & 50.0 & 50.0  & 100.0 & 66.67 & 80.0 & 60.0  \\
o3-mini& &88.89 & 77.78 &  100.0 & 44.44& 75.0 & 44.44 & 55.56& 58.93 & 30.95 & 35.71 & 64.29 & 64.29 & 64.29  & 100.0 & 77.78 & 40.0 & 40.0  \\
DeepSeek-V3& &88.89 & 77.78 &  100.0 & 44.44& 75.0 & 55.56 & 59.26& 64.88 & 33.93 & 50.0 & 67.86 & 50.0 & 50.0  & 100.0 & 33.33 & 60.0 & 20.0  \\
\textit{+CoT}& &88.89 & 77.78 &  100.0 & 22.22& 82.14 & 55.56 & 55.56& 68.45 & 45.24 & 35.71 & 75.0 & 64.29 & 64.29  & 88.89 & 88.89 & 60.0 & 60.0  \\
DeepSeek-R1& &100.0 & 66.67 &  100.0 & 11.11& 100.0 & 44.44 & 59.26& 74.4 & 40.48 & 35.71 & 82.14 & 50.0 & 50.0  & 100.0 & 55.56 & 80.0 & 40.0  \\
Qwen2.5& &100.0 & 77.78 &  100.0 & 22.22& 71.43 & 22.22 & 81.48& 70.24 & 34.52 & 21.43 & 53.57 & 42.86 & 42.86  & 100.0 & 88.89 & 40.0 & 40.0  \\
\textit{+CoT}& &88.89 & 77.78 &  100.0 & 0.0& 75.0 & 22.22 & 62.96& 72.62 & 42.26 & 0.0 & 75.0 & 35.71 & 35.71  & 100.0 & 88.89 & 20.0 & 60.0  \\
QwQ& &100.0 & 88.89 &  100.0 & 11.11& 92.86 & 22.22 & 92.59& 73.21 & 35.12 & 50.0 & 71.43 & 42.86 & 42.86  & 100.0 & 77.78 & 80.0 & 40.0  \\
Llama-3.3& &100.0 & 88.89 &  100.0 & 0.0& 64.29 & 0.0 & 88.89& 55.36 & 35.12 & 14.29 & 71.43 & 21.43 & 21.43  & 88.89 & 55.56 & 60.0 & 60.0  \\
\textit{+CoT}& &100.0 & 77.78 &  100.0 & 0.0& 67.86 & 0.0 & 77.78& 61.31 & 37.5 & 0.0 & 42.86 & 0.0 & 0.0  & 100.0 & 77.78 & 80.0 & 60.0  \\
\midrule
GPT-4o&\multirow{12}{*}{2} &85.71 & 28.57 &  0.0 & 14.29& 38.89 & 28.57 & 38.1& 56.06 & 43.18 & 66.67 & 65.0 & 72.73 & 72.73  & 100.0 & 14.29 & 50.0 & 50.0  \\
\textit{+CoT}& &85.71 & 57.14 &  0.0 & 14.29& 33.33 & 0.0 & 38.1& 59.09 & 46.97 & 22.22 & 65.0 & 9.09 & 9.09  & 85.71 & 0.0 & 50.0 & 50.0  \\
o1& &100.0 & 57.14 &  0.0 & 14.29& 38.89 & 0.0 & 66.67& 52.27 & 48.48 & 44.44 & 60.0 & 45.45 & 45.45  & 85.71 & 14.29 & 75.0 & 50.0  \\
o3-mini& &85.71 & 28.57 &  0.0 & 14.29& 44.44 & 14.29 & 33.33& 50.76 & 31.06 & 22.22 & 45.0 & 36.36 & 36.36  & 57.14 & 28.57 & 50.0 & 50.0  \\
DeepSeek-V3& &85.71 & 42.86 &  100.0 & 57.14& 55.56 & 14.29 & 33.33& 49.24 & 45.45 & 66.67 & 75.0 & 81.82 & 81.82  & 100.0 & 42.86 & 50.0 & 25.0  \\
\textit{+CoT}& &71.43 & 71.43 &  100.0 & 14.29& 50.0 & 28.57 & 47.62& 59.85 & 41.67 & 44.44 & 65.0 & 54.55 & 54.55  & 85.71 & 28.57 & 50.0 & 50.0  \\
DeepSeek-R1& &71.43 & 85.71 &  100.0 & 28.57& 72.22 & 28.57 & 61.9& 59.09 & 50.0 & 66.67 & 80.0 & 72.73 & 72.73  & 85.71 & 28.57 & 75.0 & 50.0  \\
Qwen2.5& &57.14 & 14.29 &  100.0 & 28.57& 38.89 & 0.0 & 71.43& 56.06 & 41.67 & 22.22 & 60.0 & 54.55 & 54.55  & 100.0 & 14.29 & 75.0 & 50.0  \\
\textit{+CoT}& &57.14 & 42.86 &  100.0 & 14.29& 27.78 & 0.0 & 61.9& 59.85 & 44.7 & 11.11 & 65.0 & 27.27 & 27.27  & 100.0 & 28.57 & 75.0 & 50.0  \\
QwQ& &14.29 & 100.0 &  100.0 & 42.86& 0.0 & 0.0 & 4.76& 1.52 & 22.73 & 0.0 & 5.0 & 0.0 & 0.0  & 0.0 & 14.29 & 0.0 & 0.0  \\
Llama-3.3& &100.0 & 42.86 &  100.0 & 14.29& 72.22 & 14.29 & 47.62& 41.67 & 40.15 & 22.22 & 75.0 & 45.45 & 45.45  & 85.71 & 28.57 & 75.0 & 50.0  \\
\textit{+CoT}& &100.0 & 42.86 &  100.0 & 14.29& 50.0 & 0.0 & 85.71& 53.79 & 43.18 & 0.0 & 40.0 & 0.0 & 0.0  & 100.0 & 28.57 & 50.0 & 50.0  \\
\midrule
GPT-4o&\multirow{12}{*}{3} &43.75 & 12.5 &  87.5 & 31.25& 31.58 & 25.0 & 33.33& 57.03 & 38.28 & 65.62 & 73.33 & 65.62 & 65.62  & 100.0 & 18.75 & 68.75 & 0.0  \\
\textit{+CoT}& &56.25 & 12.5 &  81.25 & 31.25& 18.42 & 0.0 & 41.67& 70.05 & 38.28 & 3.12 & 72.22 & 18.75 & 18.75  & 100.0 & 12.5 & 62.5 & 0.0  \\
o1& &25.0 & 62.5 &  81.25 & 50.0& 7.89 & 0.0 & 16.67& 23.96 & 34.9 & 12.5 & 34.44 & 15.62 & 15.62  & 50.0 & 12.5 & 37.5 & 0.0  \\
o3-mini& &6.25 & 6.25 &  93.75 & 43.75& 34.21 & 12.5 & 33.33& 53.65 & 29.69 & 43.75 & 70.0 & 53.12 & 53.12  & 100.0 & 12.5 & 56.25 & 6.25  \\
DeepSeek-V3& &43.75 & 12.5 &  93.75 & 43.75& 21.05 & 12.5 & 33.33& 56.51 & 36.46 & 78.12 & 68.89 & 68.75 & 68.75  & 100.0 & 12.5 & 56.25 & 0.0  \\
\textit{+CoT}& &50.0 & 25.0 &  93.75 & 37.5& 42.11 & 0.0 & 35.42& 64.84 & 46.35 & 40.62 & 81.11 & 65.62 & 65.62  & 100.0 & 50.0 & 50.0 & 0.0  \\
DeepSeek-R1& &56.25 & 31.25 &  81.25 & 18.75& 57.89 & 6.25 & 41.67& 64.58 & 42.71 & 65.62 & 75.56 & 62.5 & 62.5  & 100.0 & 6.25 & 75.0 & 0.0  \\
Qwen2.5& &31.25 & 12.5 &  93.75 & 31.25& 36.84 & 12.5 & 41.67& 60.42 & 36.2 & 46.88 & 65.56 & 71.88 & 71.88  & 100.0 & 12.5 & 43.75 & 0.0  \\
\textit{+CoT}& &37.5 & 31.25 &  93.75 & 37.5& 55.26 & 0.0 & 39.58& 65.1 & 37.24 & 21.88 & 71.11 & 43.75 & 43.75  & 100.0 & 87.5 & 43.75 & 6.25  \\
QwQ& &31.25 & 12.5 &  81.25 & 50.0& 55.26 & 31.25 & 45.83& 72.4 & 38.02 & 46.88 & 83.33 & 53.12 & 53.12  & 100.0 & 6.25 & 56.25 & 6.25  \\
Llama-3.3& &31.25 & 18.75 &  87.5 & 31.25& 31.58 & 0.0 & 62.5& 41.15 & 38.02 & 37.5 & 73.33 & 15.62 & 15.62  & 93.75 & 6.25 & 50.0 & 0.0  \\
\textit{+CoT}& &18.75 & 37.5 &  100.0 & 25.0& 36.84 & 0.0 & 45.83& 50.78 & 33.59 & 0.0 & 62.22 & 3.12 & 3.12  & 100.0 & 6.25 & 68.75 & 0.0  \\
\midrule
GPT-4o&\multirow{12}{*}{4} &56.25 & 6.25 &  81.25 & 31.25& 100.0 & 18.75 & 39.58& 64.2 & 47.53 & 18.52 & 72.84 & 62.96 & 62.96  & 93.75 & 62.5 & 100.0 & 18.18  \\
\textit{+CoT}& &62.5 & 12.5 &  87.5 & 43.75& 88.89 & 0.0 & 52.08& 75.93 & 54.01 & 14.81 & 82.72 & 25.93 & 25.93  & 93.75 & 68.75 & 81.82 & 36.36  \\
o1& &31.25 & 18.75 &  50.0 & 37.5& 74.07 & 6.25 & 70.83& 61.73 & 53.4 & 62.96 & 90.12 & 66.67 & 66.67  & 93.75 & 75.0 & 90.91 & 63.64  \\
o3-mini& &12.5 & 6.25 &  31.25 & 43.75& 25.93 & 0.0 & 52.08& 52.16 & 35.8 & 18.52 & 71.6 & 55.56 & 55.56  & 93.75 & 50.0 & 45.45 & 36.36  \\
DeepSeek-V3& &68.75 & 6.25 &  68.75 & 37.5& 59.26 & 12.5 & 37.5& 52.47 & 43.83 & 59.26 & 62.96 & 81.48 & 81.48  & 93.75 & 68.75 & 81.82 & 45.45  \\
\textit{+CoT}& &75.0 & 18.75 &  87.5 & 37.5& 88.89 & 18.75 & 50.0& 61.42 & 48.46 & 51.85 & 79.01 & 62.96 & 62.96  & 93.75 & 68.75 & 81.82 & 36.36  \\
DeepSeek-R1& &68.75 & 6.25 &  75.0 & 37.5& 81.48 & 18.75 & 62.5& 65.43 & 48.15 & 18.52 & 79.01 & 44.44 & 44.44  & 93.75 & 68.75 & 100.0 & 54.55  \\
Qwen2.5& &43.75 & 6.25 &  75.0 & 31.25& 70.37 & 6.25 & 52.08& 52.78 & 44.44 & 3.7 & 76.54 & 37.04 & 37.04  & 93.75 & 68.75 & 63.64 & 36.36  \\
\textit{+CoT}& &37.5 & 31.25 &  81.25 & 31.25& 66.67 & 0.0 & 60.42& 63.27 & 46.3 & 3.7 & 74.07 & 44.44 & 44.44  & 87.5 & 87.5 & 63.64 & 36.36  \\
QwQ& &12.5 & 0.0 &  31.25 & 31.25& 66.67 & 6.25 & 64.58& 73.46 & 46.6 & 25.93 & 85.19 & 37.04 & 37.04  & 87.5 & 68.75 & 81.82 & 54.55  \\
Llama-3.3& &37.5 & 18.75 &  87.5 & 31.25& 62.96 & 0.0 & 58.33& 47.22 & 51.85 & 14.81 & 76.54 & 14.81 & 14.81  & 93.75 & 62.5 & 72.73 & 54.55  \\
\textit{+CoT}& &31.25 & 12.5 &  81.25 & 37.5& 81.48 & 0.0 & 66.67& 53.4 & 45.37 & 3.7 & 70.37 & 0.0 & 0.0  & 93.75 & 68.75 & 100.0 & 54.55  \\
        \bottomrule
      \end{tabularx}
  \caption{The performance of different LLM families on our benchmark dataset.}
  \label{tab:model_performance}
\end{table*}

\subsection{Prompt Templates for Model Evaluation}
\label{appendix:prompt template}
Prompt templates used for model evaluation are listed in Table~\ref{tab:prompt_template}.

\begin{table}[!hbt]
    \centering
    \small
    \begin{tabular}{p{0.25\linewidth}p{0.65\linewidth}}
        \toprule
            \textbf{Answer Type} & \textbf{Prompt Template}   \\
        \midrule
            \textbf{Binary} & {You are an expert in social reasoning. Answer the following question with 'Yes' or 'No'. Remember: Your answer should ONLY include 'Yes' or 'No' with nothing else.\newline\# Context: \{\}\newline\# Question: \{\}\newline (Let's think step by step:)} \\
        \midrule
           \textbf{ MCQs } & {You are an expert in social reasoning. Answer the following question with the option number of the most appropriate answer. Remember: Your answer should ONLY include the option number with nothing else.\newline\# Context: \{\}\newline\# Question: \{\} \newline\# Options: \{\}\newline (Let's think step by step:)} \\
        \midrule
            \textbf{List}  & {You are an expert in social reasoning. List the required items and split them with commas. Remember: Your answer should ONLY include the required items splited by commas with nothing else.\newline\# Context: \{\}\newline\# Question: \{\}\newline(Let's think step by step:)}\\
        \midrule
            \textbf{Freeform}  & {You are an expert in social reasoning. Answer the following question with a single sentence.\newline\# Context: \{\}\newline\# Question: \{\}\newline(Let's think step by step:)}\\
        \bottomrule
    \end{tabular}
    \caption{The prompt templates for model evaluation. The CoT prompt template additionally includes the instruction ``Let's think step by step: ''.}
    \label{tab:prompt_template}
\end{table}

\subsection{ToM Method Testing}
\label{appendix:tom_methods}
In addition to CoT, there are also some ToM-specific enhancement methods~\cite{wilf-etal-2024-think, sclar-etal-2023-minding, zhang2025autotomscalingmodelbasedmental, shinoda2025letssallysshoesshoesofothers}. But these methods most were designed for Sally-Anne-style belief-tracking tasks and cannot be directly applied to our white lie scenarios due to three key challenges:

\begin{itemize}
\item \textbf{Multiple asymmetries}: Each white lie scenario involves an information triplet (truth, lie, real reason), requiring models to track multiple asymmetries simultaneously, unlike Sally-Anne tasks that focus on a single hidden fact.

\item \textbf{Graduated information access}: Different characters have varying levels of access to triplet information, contrasting with the binary "knows/doesn't know" structure in traditional ToM tasks.

\item \textbf{Complex reasoning}: Our questions extend beyond simple belief tracking to evaluate models' ability to apply multi-role beliefs in white lie reasoning (e.g., comprehension and justification questions).
\end{itemize}

Thus, direct application would require substantial redesign. We tested the most applicable approach, the \textit{Let's Put Ourselves in Sally's Shoes} method~\cite{shinoda2025letssallysshoesshoesofothers}, which makes fewer assumptions about context and suits Belief and LieAbility questions focusing on individual perspectives. 
As shown in Table \ref{tab:gpt4o_results}, the marginal or inconsistent improvements (+2.38\% for Belief but -1.26\% for LieAbility) suggest that single-person perspective-taking approaches designed for belief-tracking tasks need substantial adaptation for white-lie scenarios. These results reinforce our analysis of the fundamental differences between belief-tracking and white-lie comprehension, highlighting the need for specialized methods to handle the complex information structure and multi-role beliefs inherent in white-lie scenarios.

\begin{table}[h]
\centering
\resizebox{0.8\columnwidth}{!}{%
\begin{tabular}{lcc}
\toprule
\textbf{Question} & 
\textbf{\begin{tabular}[c]{@{}c@{}}Standard \\ GPT-4o\end{tabular}} & 
\textbf{\begin{tabular}[c]{@{}c@{}}GPT-4o with \\ SoO\end{tabular}} \\
\midrule
Prefixing Belief & 61.56 & 63.94 \\
LieAbility & 60.62 & 59.36 \\
\bottomrule
\end{tabular}%
}
\caption{Results (\%) by the \textit{Let's Put Ourselves in Sally's Shoes} method \cite{shinoda2025letssallysshoesshoesofothers}.}
\label{tab:gpt4o_results}
\end{table}

\subsection{More Analysis}
\label{appendix:more_analysis}
\begin{figure}[t]
    \centering
    \includegraphics[width=1\linewidth]{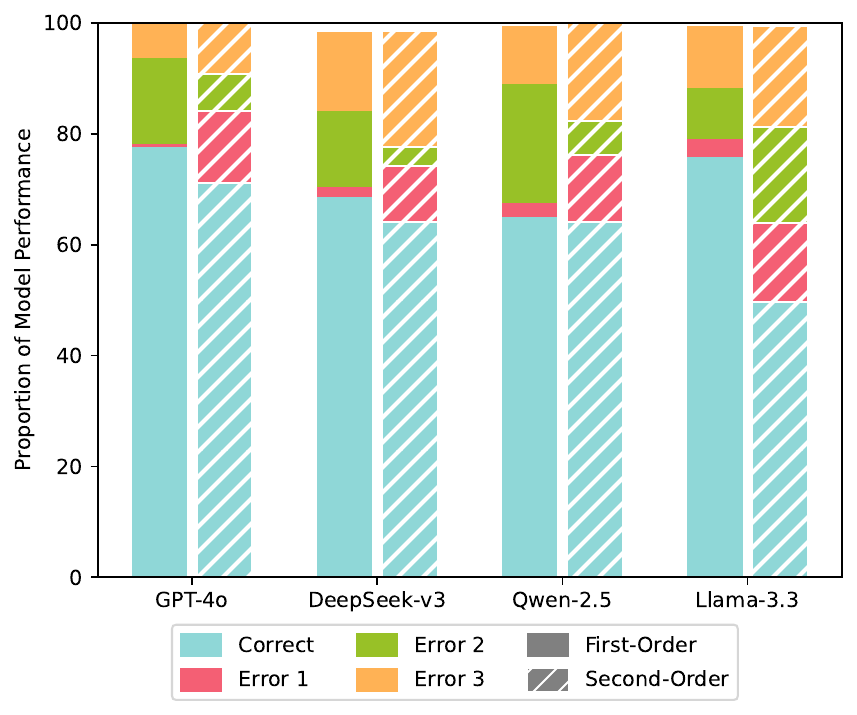}
    \caption{The proportion of model performance types in BeliefQA questions.}
    \label{fig:error_proportion_belief}
\end{figure}

\begin{figure}[t]
    \centering
    \includegraphics[width=1\linewidth]{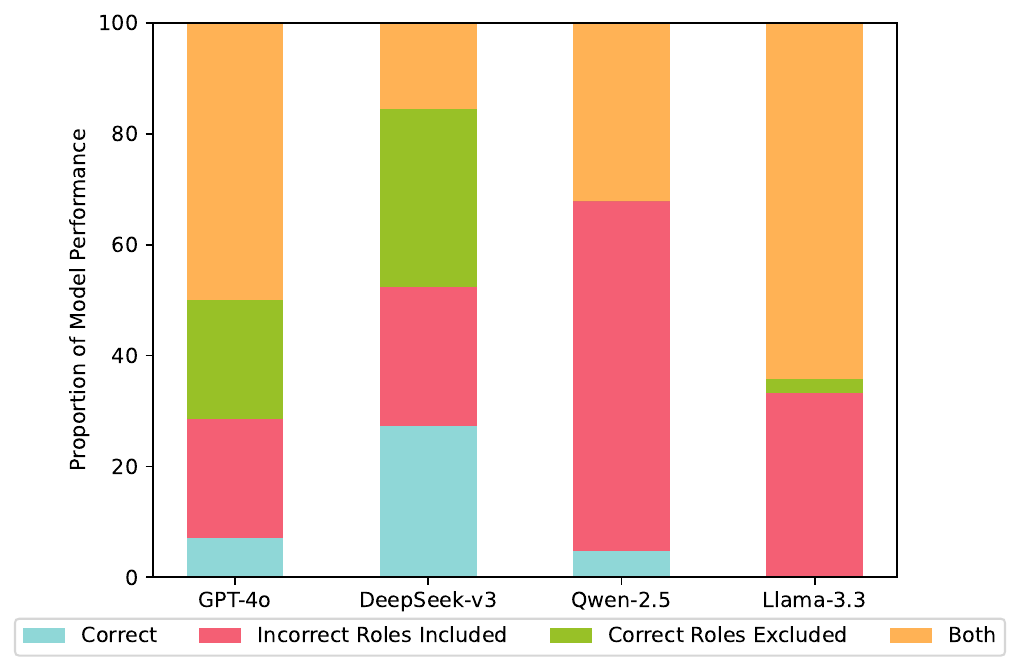}
    \caption{The proportion of model performance types in Lie Detectability questions of list format.}
    \label{fig:error_proportion_liedetectablity}
\end{figure}

We provide additional error analysis by tracking the specific wrong options selected by models across different question types, which provide further insights into the failure modes of various LLMs when handling white lie scenarios. The distribution of error types for Belief Understanding (Figure \ref{fig:error_proportion_belief}), and Role-Specific Performance in Lie Detection (Figure \ref{fig:error_proportion_liedetectablity}) reveal systematic patterns in how models misunderstand white lie contexts.

\section{AI Usage and Resources}
In this project, we used LLMs for assistance. 
During paper writing, we used models from GPT and Claude families to help us refine and enhance our expressions. For programming, we also relied on models from GPT family to generate reference code, which we subsequently modified to complete our tasks.

Icons used in Figure~\ref{fig:example} and Figure~\ref{fig:pipeline} were generated using Recraft\footnote{\url{https://www.recraft.ai/}} and GPT-4o~\cite{hurst2024gpt} or sourced from Icons8\footnote{\url{https://icons8.com/icons}}.

\typeout{get arXiv to do 4 passes: Label(s) may have changed. Rerun}
\end{document}